\documentclass[10pt,letterpaper,twocolumn]{article}
\usepackage[latin9]{inputenc}
\usepackage{array}
\usepackage{textcomp}
\usepackage{mathtools}
\usepackage{multirow}
\usepackage{amsmath}
\usepackage{amssymb}
\usepackage[unicode=true,
 bookmarks=false,
 breaklinks=true,pdfborder={0 0 1},backref=section,colorlinks=false]
 {hyperref}
\hypersetup{
 pagebackref=true,letterpaper=true,colorlinks}

\makeatletter

\pdfpageheight\paperheight
\pdfpagewidth\paperwidth

\newcommand{\lyxmathsym}[1]{\ifmmode\begingroup\def\b@ld{bold}
  \text{\ifx\math@version\b@ld\bfseries\fi#1}\endgroup\else#1\fi}

\providecommand{\tabularnewline}{\\}

\usepackage{iccv}
\usepackage{times}
\usepackage{epsfig}
\usepackage{graphicx}
\usepackage{import}

\iccvfinalcopy 

\makeatother

\begin{document}

\title{Self-Supervised Reversed Image Signal Processing via Reference-Guided
Dynamic Parameter Selection}
\author{Junji Otsuka\quad{}Masakazu Yoshimura\quad{}Takeshi Ohashi\\
 Sony Group Corporation\\
 \texttt{\small{}\{junji.otsuka, masakazu.yoshimura, takeshi.a.ohashi\}@sony.com}}

\maketitle


\begin{abstract}
Unprocessed sensor outputs (RAW images) potentially improve both low-level
and high-level computer vision algorithms, but the lack of large-scale
RAW image datasets is a barrier to research. Thus, reversed Image
Signal Processing (ISP) which converts existing RGB images into RAW
images has been studied. However, most existing methods require camera-specific
metadata or paired RGB and RAW images to model the conversion, and
they are not always available. In addition, there are issues in handling
diverse ISPs and recovering global illumination. To tackle these limitations,
we propose a self-supervised reversed ISP method that does not require
metadata and paired images. The proposed method converts a RGB image
into a RAW-like image taken in the same environment with the same
sensor as a reference RAW image by dynamically selecting parameters
of the reversed ISP pipeline based on the reference RAW image. The
parameter selection is trained via pseudo paired data created from
unpaired RGB and RAW images. We show that the proposed method is able
to learn various reversed ISPs with comparable accuracy to other state-of-the-art
supervised methods and convert unknown RGB images from COCO and Flickr1M
to target RAW-like images more accurately in terms of pixel distribution.
We also demonstrate that our generated RAW images improve performance
on real RAW image object detection task.
\end{abstract}

\section{Introduction}

In general, a sensor RAW image taken by a digital camera is converted
into the standard RGB (sRGB) format through an in-camera ISP \cite{karaimer2016software}.
Traditional ISPs are essentially optimized to generate compressed
and human perceptually pleasant RGB images. Due to the ease of use,
numerous RGB images flood on the Internet, and their availability
underpin recent advance in machine learning-based computer vision
technologies. On the other hand, RAW images contain all the captured
information, and the relationship between ambient light, pixel intensity,
and noise distribution in RAW domain is much simpler than that in
RGB domain \cite{yoshimura2022rawgment}. Therefore, utilizing RAW
images directly for downstream tasks potentially achieves greater
performance than RGB image-based methods in both low-level and high-level
computer vision tasks. In fact, recent studies have shown that RAW
image-based image recognition \cite{morawski2022genisp,Hong2021Crafting,yoshimura2022dynamicisp}
and image processing \cite{brooks2019unprocessing,zamir2020cycleisp,zhang2019zoom,liang2020raw}
achieved higher performance than RGB image-based methods. The use
of RAW images is expected to improve performance especially in difficult
scenes such as extremely dark or blurry scenes that should be covered
in practical application. RAW images are also used in research that
optimizes existing ISPs for downstream tasks \cite{pavithra2021automatic,mosleh2020hardware,hevia2020optimization,nishimura2018automatic,tseng2019hyperparameter,robidoux2021end,yu2021reconfigisp}
or develops an accurate DNN-based ISP \cite{liu2022deep,schwartz2018deepisp,ignatov2020replacing}.
However, they are hard-to-see and not suitable for daily use. Therefore,
it is difficult to obtain enough RAW images for research purposes.
In particular, the scarcity of annotated RAW data has been a barrier
to machine learning-based approaches. Hence, several reversed ISP
methods that convert existing large-scale RGB datasets into pseudo
RAW datasets have been studied \cite{brooks2019unprocessing,zamir2020cycleisp,conde2022model,xing21invertible,conde2022aim,kim2012new,afifi2021cie}.

The reversed ISP methods can be divided into model-based methods \cite{kim2012new,brooks2019unprocessing}
and learning-based methods \cite{zamir2020cycleisp,conde2022model,xing21invertible,afifi2021cie,conde2022aim}.
UPI \cite{brooks2019unprocessing} is a typical model-based method
that defines a series of simple and invertible ISP blocks whose parameters
are determined using camera metadata such as white balance gains and
color correction matrixes. On the other hand, learning-based methods
learn RGB-to-RAW conversion directly from paired RGB and RAW images
using a fully DNN based model \cite{zamir2020cycleisp,afifi2021cie,xing21invertible,conde2022aim}
or a combination of hand-crafted ISP blocks and shallow CNN models
\cite{conde2022model,conde2022aim}. Learning-based methods are able
to achieve more accurate RAW reconstruction than model-based methods.

These methods are pioneers of reversed ISP research and valuable to
overcome the shortage of RAW images. However, there are some limitations.
First, camera metadata or paired RAW and RGB data is required. Metadata
is often not accessible for camera users, and shooting RGB and RAW
images simultaneously is not executable in all cases. Second, most
existing methods assume a single specific ISP pipeline and do not
handle RGB images processed by unknown ISP well. This causes misaligned
color and brightness distribution when applied to arbitrary RGB datasets.
Finally, it is hard to reproduce characteristics of global illumination
of a target RAW image from an input RGB image since the effect of
global illumination is generally canceled in RGB images by ISPs. Therefore,
the existing methods tend to produce different RAW image distribution
from the target distribution.

To tackle these problems, in this paper, we present an Self-supervised
Reversed ISP method called SRISP that does not require metadata and
paired data for training. Same as MBISPLD \cite{conde2022model},
SRISP has multiple parameter dictionaries of a reversed ISP pipeline
composed of classical ISP functions and shallow CNNs to cover various
kinds of sensors and environments. Then, the parameters are selected
by other shallow CNNs to achieve correct mapping. Unlike MBISPLD,
the proposed method (1) only needs source RGB images and unpaired
target RAW images to train it with the help of proposed two types
of pseudo image pairs and (2) achieves diverse RGB-to-RAW mapping
including illumination effects by conditioning the selection module
with global features of reference target RAW and source RGB images.
Our main contributions are: \vspace{-2mm}
\begin{itemize}
\item Self-supervised reversed ISP learning based on two types of pseudo
paired data generated from unpaired target RAW and source RGB images
using a randomized traditional ISP and novel self-supervision based
on Mean Teacher \cite{tarvainen2017mean}. \vspace{-2mm}
\item Dynamic parameter selection of reversed ISP blocks using global features
of a reference RAW image, which is able to reproduce the target RAW
characteristic including global illumination. \vspace{-2mm}
\item Demonstrate that our method is able to map existing RGB datasets (COCO
\cite{lin2014microsoft} and Flickr1M \cite{huiskes08}) to target
RAW datasets (MIT-Adobe FiveK \cite{fivek}, SIDD \cite{sidd}, and
LOD Dataset \cite{Hong2021Crafting}) more accurately than other state-of-the-art
methods in terms of pixel distribution. \vspace{-2mm}
\item Additional experiments that show our learned model contributes to
the accuracy improvement in RAW object detection on LOD Dataset.
\end{itemize}

\section{Relate Work}

\subsection{Reversed ISP}

The learning-based methods are further divided into fully DNN-based
methods \cite{zamir2020cycleisp,afifi2021cie,xing21invertible,conde2022aim,kinli2023reversing,dong2023rispnet,zou2023learned,kim2023overexposure}
and hybrid-methods \cite{conde2022model,conde2022aim}. CycleISP \cite{zamir2020cycleisp}
and InvISP \cite{xing21invertible} are state-of-the-art fully DNN-based
methods. CycleISP models RGB-to-RAW and RAW-to-RGB mappings using
two DNN branches that are jointly fine-tuned to achieve cycle consistency.
InvISP learns a reversible ISP using normalizing flow \cite{kingma2018glow,ho2019flow++}
to produce an invertible RGB image to the original RAW image. While
the fully DNN-based methods are expressive, there are issues of interpretability
and controllability. On the other hand, MBISPLD \cite{conde2022model}
employs a hybrid approach combining UPI-like classical reversible
ISP blocks and shallow CNNs. Each ISP block has learnable parameters
optimized with RGB and RAW image pairs. As for the white balance and
color corretion blocks, multiple candidate parameters (parameter dictionary)
are learned and dynamically selected by shallow CNNs based on intermediate
images at inference time. They also use shallow CNNs as the learnable
lens shading correction and tone mapping. MBISPLD achieved state-of-the-art
RAW reconstruction accuracy while maintaining the interpretability.
Our method is an extension of MBISPLD, which is composed of classic
ISP blocks and shallow CNNs. The main differences are the parameter
selection module and the self-supervised learning method. Our method
selects optimal parameters of ISP blocks based on global features
of source and reference images and is end-to-end trainable with unpaired
RGB and RAW images.

\subsection{ISP Optimization and Control}

Recently, several methods have been proposed to optimize parameters
of a classic ISP to improve the performance of downstream tasks or
perceptual image quality \cite{pavithra2021automatic,mosleh2020hardware,hevia2020optimization,nishimura2018automatic,tseng2019hyperparameter,robidoux2021end,yu2021reconfigisp}.
For example, in \cite{tseng2019hyperparameter}, the differentiable
proxy that mimics the behavior of a non-differentiable ISP function
using DNN is trained, and the ISP parameters are optimized based on
the proxy directly using gradient descent to maximize performance
of several downstream tasks. Similarly, Covariance Matrix Adaptation
Evolution Strategy \cite{hansen1996adapting} is used to optimize
black-box ISP parameters \cite{mosleh2020hardware}. In addition,
ReconfigISP \cite{buckler2017reconfiguring} optimizes both the combination
of ISP blocks and their parameters with a neural architecture search
method \cite{liu2018darts}. Unlike these static ISP optimization,
several methods \cite{onzon2021neural,yoshimura2022dynamicisp,morawski2022genisp,Cui_2022_BMVC}
dynamically control ISP parameters such as the digital gain, white
balance, denoiser, and tone mapping to enhance the downstream performance.
These studies show that even a model-based ISP with limited expressiveness
achieves high performance by optimizing or controlling its parameters.

\subsection{Pseudo Labeling}

In the field of image recognition, pseudo-labeling that treats outputs
of a specific model (teacher) as ground truth data is widely used
when the real ground truth data is not available or noisy. In particular,
Mean Teacher (MT) \cite{tarvainen2017mean}, which uses an Exponential
Moving Average (EMA) model of past training step models as a teacher,
has been shown to be effective in self-supervised learning, semi-supervised
learning, and domain adaptation in recent years \cite{li2021synthetic,li2022cross,liu2021unbiased,Liu_2022_CVPR}.
MT supresses the error of the pseudo-labels by the temporal model
ansamble and is expected to achieve stable training. In this study,
we propose a MT-based pseudo-paired data generation method for better
training.

\section{Method}

\begin{figure*}
\centering

\def\svgwidth{2.0\columnwidth}

\scriptsize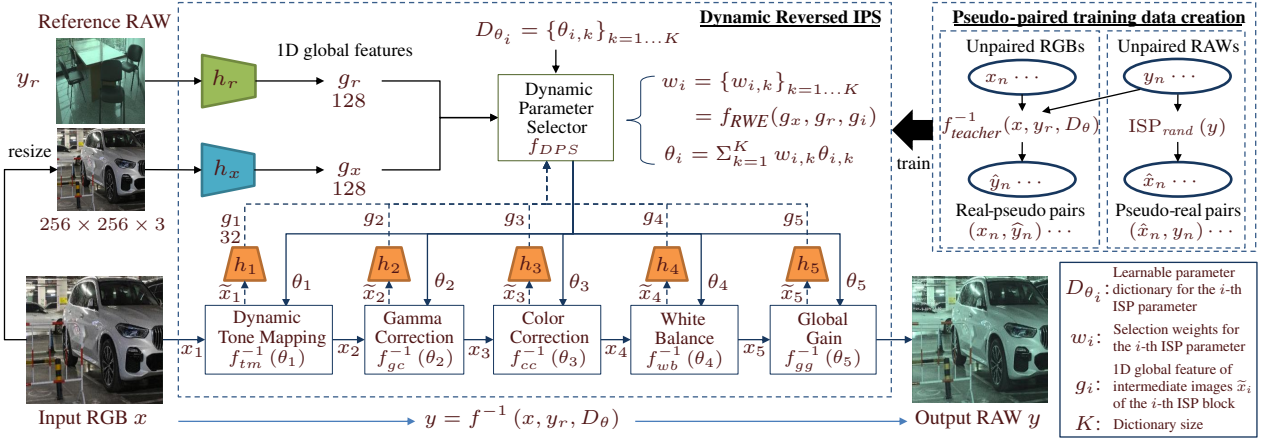

\caption{Our SRISP framework. An input RGB image and a reference RAW image
are converted into 1D global features, and ISP parameters are dynamically
determined based on them. To model complex mappings, a new reference-based
dynamic CNN is incorporated as Dynamic Tone Mapping. Two types of
psedo-paired RGB and RAW images are used for stable training.}

\label{fig:model_overview}
\end{figure*}

Figure \ref{fig:model_overview} shows an overview of our proposed
approach. Let us denote $\mathcal{X}$ as RGB image domain and $\mathcal{Y}$
as RAW image domain. Our goal is to find the reversed mapping $f^{-1}\vcentcolon\mathcal{X\rightarrow\mathcal{Y}}$.
Similar to UPI \cite{brooks2019unprocessing} and MBISPLD \cite{conde2022model},
we modeled the mapping function by a series of differentiable and
reversible ISP blocks: Global Gain (GG) $f_{gg}$, White Balance (WB)
$f_{wb}$, Color Correction (CC) $f_{cc}$, Gamma Correction (GC)
$f_{gc}$, and Tone Mapping (TM) $f_{tm}$. Note that bilinear demosicing
is applied before $f_{gg}$ as preprocessing because it is deterministic
processing, and the demosaiced image is treated as a RAW image in
this paper. The RGB-to-RAW mapping $f^{-1}$ is defined as follows: 

\begin{equation}
f^{-1}=f_{gg}^{-1}\circ f_{wb}^{-1}\circ f_{cc}^{-1}\circ f_{gc}^{-1}\circ f_{tm}^{-1}.
\end{equation}
The $i$-th ISP block has a parameter dictionary $D_{\theta_{i}}$
with $K$ parameter candidates $\left\{ \theta_{i,k}\right\} _{k=1...K}$.
The final parameter $\theta_{i}$ of the $i$-th block is determined
by a Dynamic Parameter Selector (DPS) $f_{DPS}$ based on the parameter
dictionary so that a given RGB image $x$ is converted into a corresponding
RAW image $y$. However, in general, if the functions and parameters
of the forward mapping are unknown, the inversion estimation is ill-posed
as there can be many RAW images corresponding to an input RGB image.
It is hard to estimate what true illumination was and how the RGB
image was processed from only the input RGB image since thier clues
are essentially removed by a forward ISP. Therefore, we solve this
problem by giving a reference RAW image $y_{r}$ to the DPS. Then,
the proposed method is able to convert \textbf{$x$} processed by
an arbitrary ISP into a $y_{r}$-like RAW image \textbf{$y$}. Fothermore,
the parameter dictionaries and the DPS are tranined using unpaired
RGB and RAW images.

We implemented GG, WB, CC, and GC based on UPI. Due to space limitation,
we detail them in the supplementary material. The key point is, while
MBISPLD unifies CG into WB with a single parameter dictionary, we
designed them separately with independent dictionaries to enhance
their flexibility. As for TM, we introduced new Dynamic Tone Mapping
(DTM) described in Section 3.2.

\subsection{Dynamic Parameter Selector}

\begin{figure}
\centering

\def\svgwidth{1.0\columnwidth}

\scriptsize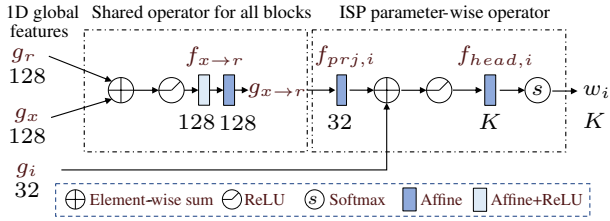

\caption{Reference-guided Weight Estimator $f_{\ensuremath{RWE}}$ determines
ISP parameter selection weights based on 1D global features calculated
from an input image, a reference image, and intermidiate features
of the $i$-th ISP block.}

\label{fig:DPS}
\end{figure}

Following MBISPLD, we designed our DPS to estimate the parameter $\theta_{i}$
of the $i$-th ISP block as a weighted average of the parameter dictionary
$D_{\theta_{i}}$ as follows:

\begin{equation}
\theta_{i}=\mathop{\Sigma_{k=1}^{K}}w_{i,k}\theta_{i,k}.\label{eq:dps1}
\end{equation}
Unlike MBISPLD, our model attempts to map an input RGB image $x$
to a RAW-like image $y$ similar to reference RAW image $y_{r}$.
To this end, the weights $w_{i}=\left\{ w_{i,k}\right\} _{k=1...K}$
are determined by Reference-guided Weight Estimator (RWE) $f_{\ensuremath{RWE}}$
based on 1D global features $g_{x}=h_{x}\left(x\right)$, $g_{r}=h_{r}\left(y_{r}\right)$,
and $g_{i}=h_{i}\left(\widetilde{x}_{i}\right)$:

\begin{equation}
w_{i}=f_{RWE}\left(g_{x},g_{r},g_{i}\right),\label{eq:dps2}
\end{equation}
where $h_{x}$, $h_{r}$, and $h_{i}$ are shallow CNNs followed by
the global average pooling, and $\widetilde{x}_{i}$ is intermidiate
features of the $i$-th ISP block. For the blocks except for the DTM,
following MBISPLD, $\widetilde{x}_{i}$ is the concatenated output
image processed by the ISP block using the $K$ parameter candidates.
As for the DTM, the input image $x$ is used as $\widetilde{x}_{i}$
to avoid heavy computation. As shown in Figure \ref{fig:DPS}, the
$f_{\ensuremath{RWE}}$ simply fuses the three global features using
affine layers ($f_{x\rightarrow r}$, $f_{\ensuremath{prj,i}}$, and
$f_{head,i}$) and element-wise sum operations, and outputs $w_{i}$
with a softmax operation. The $f_{x\rightarrow r}$ is shared for
all ISP blocks and generates the fused global feature $g_{x\rightarrow r}$
of $g_{x}$ and $g_{r}$. The $g_{x\rightarrow r}$ is expected to
represent general information how to convert $x$ to $y_{r}$. The
$f_{\ensuremath{prj,i}}$ and $f_{head,i}$ are trained for each parameter
dictionary and select suitable parameters based on $g_{x\rightarrow r}$
and $g_{i}$. Thanks to $g_{x\rightarrow r}$, our method is able
to reproduce features of the reference RAW image and generate diverse
pseudo RAW images using $y_{r}$ randomly sampled from the target
RAW data. Note that a full-resolution image is firstly resized to
$256\times256$, and ISP parameters are determined based on the resized
image to reduce a computational cost. Then, the full-resolution image
is processed using the parameters.

\subsection{Dynamic Tone Mapping}

The tone mapping maps one color to another color to render a perceptually
pleasant image, and the mapping function can be highly complex. To
model the complex mapping, we employed a shallow CNN as TM similar
to MBISPLD: 

\begin{equation}
f_{tm}\left(x,\theta_{1}\right)=\phi_{tm}\left(x,\left\{ \theta_{tm,l}\right\} _{l=1...4}\right),
\end{equation}
where $\phi_{tm}$ is a 4-layer CNN with 32 channel dimension and
the $\theta_{tm,l}$ weights for the $l$-th layer. In MBISPLD, the
$\phi_{tm}$ consists of only $1\times1$ convolutions and statically
optimized. This limited structure helps to stabilize training, but
we extended it to cover more diverse mapping. The proposed DTM uses
$3\times3$ dynamic convolutions \cite{yang2019condconv,chen2020dynamicc}
whose kernels are dynamically determined by the DPS. In the DTM, the
parameter dictionary $D_{\theta_{tm}}$ is learned for each $\theta_{tm,l}$.
That is, $D_{\theta_{1}}=\left\{ D_{\theta_{tm,l}}\right\} _{l=1...4}$
and each $D_{\theta_{tm,l}}$ includes $K$ candidate parameters for
the $l$-th layer. The $\theta_{tm,l}$ is determined as follows:

\begin{equation}
\theta_{tm,l}=f_{DPS}\left(g_{x},g_{r},g_{1},D_{\theta_{tm,l}}\right),
\end{equation}
where $f_{DPS}$ expresses (\ref{eq:dps1}) and (\ref{eq:dps2}).
The DTM has different parameter dictionaries for forward and reversed
passes since this function is not invertible. Note that the original
dynamic convolution selects its parameter based on input features
of each convolution. On the other hand, our DTM determines the weights
using global features of the input RGB and the reference RAW images.

\subsection{Pseudo Pair Training}

The proposed self-supervised learning using unpaired RGB and RAW images
is realized by combining two types of pseudo RGB and RAW image pairs
and our reference-guided DPS. The two types of pseudo-pairs are (1)
$\mathrm{PP}_{rand}$: the real-RAW $y$ and pseudo-RGB $\hat{x}$
pair generated by a random ISP and (2) $\mathrm{PP_{\mathit{MT}}}$:
the real-RGB $x$ and pseudo-RAW $\hat{y}$ pair generated by self-supervision
based on Mean Teacher (MT) \cite{tarvainen2017mean}. These pseudo-pairs
do not represent the correspondence between real RGB and RAW images.
Therefore, to learn fixed reversed pipeline using these pairs results
in poor generalization for real RGB and RAW pairs. Moreover, MT converges
to a meaningless solution when there is no correct label because MT
basically just reduces noises of the labels. However, by combined
with the reference-guided DPS, these pseudo-pairs act as correct supervision.
That is, in terms of dynamic parameter selection, our model is able
to learn how to determine the mapping parameters for the given pair
based on the reference, even if the pair is not a true pair. Furthermore,
the learned parameter selection works with an unknown true RGB and
RAW image pair. 

\begin{figure}
\centering

\def\svgwidth{0.9\columnwidth}

\scriptsize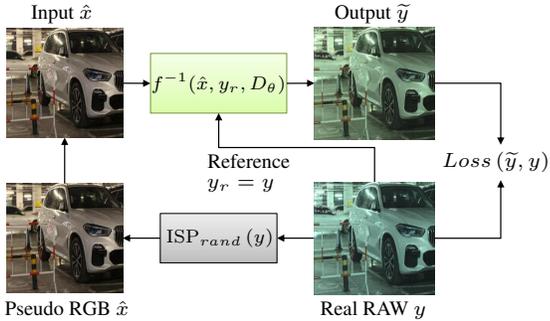

\caption{Randomized traditional ISP for creating the real RAW and pseudo RGB
pair for training.}

\label{fig:RandISP}
\end{figure}

The first pseudo pair, $\mathrm{PP}_{rand}$, is defined as:

\begin{equation}
\left(\hat{x},y\right)=\left(\mathrm{ISP}_{\mathrm{\mathit{rand}}}\left(y\right),y\right),
\end{equation}
where $\mathrm{ISP}_{\mathrm{\mathit{rand}}}$ is a simple forward
ISP pipeline same as UPI \cite{brooks2019unprocessing}. Unlike UPI,
the parameters of GG, CC, and GM are randomly determined independent
of the target mapping, and a simple Gray-world algorithm \cite{ebner2007color}
is used as WB. The implementation details are in the supplementary
material. Although $\mathrm{ISP}_{rand}$ is different from real camera
ISP, it is able to produce perceptually acceptable RGB images. By
learning to reproduce the real RAW image $y$ from this pseudo-RGB
image $\hat{x}$ as show in Figure \ref{fig:RandISP}, the proposed
method learns the basic procedure of the RGB-to-RAW mapping. Unfortunately,
RGB images generated by $\mathrm{ISP_{\mathit{rand}}}$ do not cover
true diverse distribution of RGB images and it may degrade RAW reconstruction
quality when the input is real RGB image. Therefore, the second pseudo-pair
is needed.

\begin{figure}
\centering

\def\svgwidth{0.9\columnwidth}

\scriptsize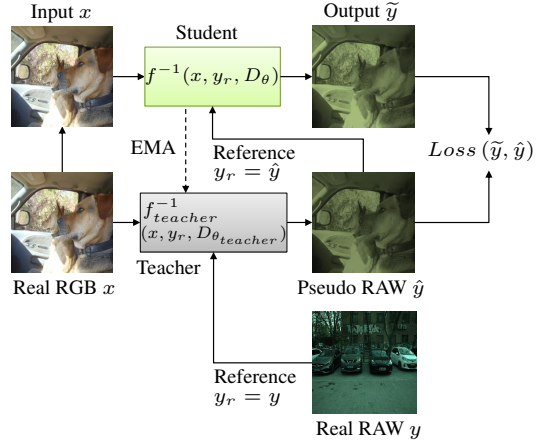

\caption{Mean Teacher model for creating the real RGB and pseudo RAW pair for
training.}

\label{fig:MT}
\end{figure}

The second pseudo-pair, $\mathrm{PP_{\mathit{MT}}}$, is defined as:

\begin{equation}
\left(x,\hat{y}\right)=\left(x,f_{teacher}^{-1}\left(x,y_{r},D_{\theta_{teacher}}\right)\right),
\end{equation}
where $f_{teacher}^{-1}$ and $D_{\theta_{teacher}}$ are MTs of $f^{-1}$
and $D_{\theta}$, respectively. We input a real RGB image $x$ and
reference real RAW image $y_{r}$ randomly sampled from the RAW dataset
into the MT and obtain an output RAW-like image $\hat{y}$. Note that
$x$ and $y_{r}$ are unpaired. Figure \ref{fig:MT} shows how to
train the model using this pseudo pair. The key point here is to give
$\hat{y}$ as the reference $y_{r}$ to the student model. This enables
the student model to learn the RGB-to-RAW mapping without any contradiction
even when the MT produces unnatural pseudo RAW images. On the other
hand, if the real RAW image $y$ set to $y_{r}$ and $\hat{y}$ is
used only for student loss calculation as in the standard MT, the
mismatch between $y_{r}$ and the target $\hat{y}$ leads to a wrong
mapping. Note that this self-supervised learning is necessary to be
conbined with the $\mathrm{PP}_{rand}$ because the MT itself does
not have a power to get close to the true mapping. The $\mathrm{PP}_{rand}$
enables the model to learn how to perform basic conversions to true
RAW images, and the $\mathrm{PP_{\mathit{MT}}}$ enables the model
to learn conversions from a variety of true RGB images. By combining
these two with the reference-guided DPS, the proposed method achieves
the un-paired learning.

\subsection{Losses}

Our goal is to build reversed ISP pipeline $f^{-1}$ with its forward
pipeline $f$ is also trained as a constraint. Hence, the following
bi-directional loss function is used: 
\begin{equation}
L_{bi}\left(x,y\right)=\left|f_{gc}\left(y\right)-f_{gc}\left(f^{-1}\left(x\right)\right)\right|+\left|x-f\left(y\right)\right|,
\end{equation}
where $f_{gc}$ is a gamma transformation with $\gamma=2.2^{-1}$
introduced to encourage learning of dark areas as well as bright area.
A similar idea was employed in \cite{zamir2020cycleisp}. We first
process $f^{-1}$, and the same weights for the parameter selection
are used for $f$. The final loss is a weighted sum of the losses
for the first and second pseudo-pairs:

\begin{equation}
L=L_{bi}\left(\hat{x},y\right)+\alpha L_{bi}\left(x,\hat{y}\right),
\end{equation}
where $\alpha$ is a weight parameter and is set to 0.3 in this paper.
The second term is not used in the first 15 epochs because outputs
of the MT are not very meaningful. 

\section{Experiments}

\subsection{Datasets}

We evaluated our method on three RAW image datasets and two RGB image
datasets widely used in verious research including high-level computer
vision studies. \textbf{MIT-Adobe-FiveK Dataset} \cite{fivek}. We
used the same train and test set used in \cite{xing21invertible}
for Canon EOS 5D (777 images) and Nikon D700 (590 images). The LibRaw
library was used to render RGB images from RAW images. \textbf{SIDD}
\cite{sidd}. This dataset provides 320 RGB and RAW image pairs captured
by five smartphone cameras under different lighting conditons for
training and 1280 patches for validation. \textbf{LOD Dataset} \cite{Hong2021Crafting}.
LOD Dataset contains low-light and normal-light RAW image pairs of
eight object categories. We used normal-light images with 1830 training
and 400 test images. We converted the original images to the DNG format
by the Adobe DNG Converter and used full-size thumbnail images as
RGB images. \textbf{COCO} \textbf{Dataset} \cite{lin2014microsoft}.
Large-scale real world object images in RGB format are provided. We
used 2017 Train images (118K images) for the $\mathrm{PP_{\mathit{MT}}}$
training and 2017 Val images (5K) for unpaired evaluation. \textbf{Flickr
1 Million Dataset} \cite{huiskes08}. This dataset also provides diverse
real world RGB images. We randomly sampled 200K images for the $\mathrm{PP_{\mathit{MT}}}$
training and 5K images for unpaired evaluation from 1M images.

\begin{table*}
\begin{center}\scalebox{0.9}{

\begin{tabular}{c|l|c|c|c|c|c|c|c|c}
\hline 
\multirow{2}{*}{Train} & \multirow{2}{*}{Method} & \multicolumn{2}{c|}{Nikon D700} & \multicolumn{2}{c|}{Canon EOS 5D} & \multicolumn{2}{c|}{SIDD} & \multicolumn{2}{c}{LOD}\tabularnewline
\cline{3-10} \cline{4-10} \cline{5-10} \cline{6-10} \cline{7-10} \cline{8-10} \cline{9-10} \cline{10-10} 
 &  & AE$\downarrow$ & PSNR$\uparrow$ & AE$\downarrow$ & PSNR$\uparrow$ & AE$\downarrow$ & PSNR$\uparrow$ & AE$\downarrow$ & PSNR$\uparrow$\tabularnewline
\hline 
\multicolumn{1}{c}{} & UPI \cite{brooks2019unprocessing} & 7.80 & 27.93 & 7.35 & 32.97 & 8.82 & 36.29 & 8.40 & 30.69\tabularnewline
\multicolumn{1}{c}{} & CycleISP \cite{zamir2020cycleisp} & 8.80 & 30.51 & 9.80 & 32.14 & 9.49 & 42.07 & 10.35 & 22.33\tabularnewline
\hline 
 & U-Net \cite{conde2022aim} & 4.83 & 38.01 & 4.73 & 41.53 & 7.76 & 45.44 & 6.06 & 38.94\tabularnewline
 & MBISPLD \cite{conde2022model} & 4.72 & 38.49 & 4.72 & 41.38 & 7.60 & 45.56 & 6.22 & 37.74\tabularnewline
Real & Ours w/o $\mathrm{PP_{\mathit{MT}}}$ & \textbf{2.53} & 43.05 & \textbf{2.84} & \textbf{45.92} & \textbf{4.77} & \textbf{49.20} & \textbf{4.62} & \textbf{40.20}\tabularnewline
 & Ours (Flickr) & 2.56 & 43.13 & 2.93 & 45.54 & 4.92 & 48.72 & 4.64 & \textbf{40.20}\tabularnewline
 & Ours (COCO) & 2.57 & \textbf{43.32} & 2.90 & 45.64 & 4.96 & 48.85 & 4.64 & 39.66\tabularnewline
\hline 
\multirow{4}{*}{Pseudo} & Ours w/o $\mathrm{PP_{\mathit{MT}}}$ & 4.55 & 34.67 & 4.00 & 37.89 & 6.86 & 42.79 & 5.05 & 35.17\tabularnewline
 & Ours (Flickr) & 3.81 & 35.51 & 3.72 & 37.97 & 5.84 & 43.68 & 4.91 & 34.64\tabularnewline
 & Ours (COCO) & 3.61 & 35.52 & 3.59 & 38.36 & 6.00 & 43.51 & 4.90 & 34.96\tabularnewline
 & Ours (All) & 3.02 & 38.80 & 3.20 & 41.18 & 5.23 & 46.23 & 4.91 & 34.89\tabularnewline
\hline 
\end{tabular}}\end{center}

\caption{Quantitative RAW reconstuction results among our methods and other
methods. The characters in parentheses of ours denotes the dataset
used for the $\mathrm{PP_{\mathit{MT}}}$. \textquotedblleft Real\textquotedblright{}
and \textquotedblleft Pseudo\textquotedblright{} indicate whether the
real paired data or the $\mathrm{PP}_{rand}$ was used for training.}

\label{tab:known_ISP}
\end{table*}


\subsection{Implementation Details}

We used the following settings for all experiments. The image encoder
$h_{r}$ and $h_{x}$ were 5-layer CNNs with $3\times3$ convolutions
of stride 2 followed by ReLU activation except for the last layer.
The channel sizes were \{32, 64, 128, 128, 128\}. The $h_{i}$ had
slightly different structure, i.e., 4-layer CNN with \{32, 32, 32,
32\} channel size. The global average pooling was applied to the outputs
of $h_{r}$, $h_{x}$, and $h_{i}$. The parameter dictionary size
$K$ was 5. The length of global feature vector $g_{r}$, $g_{x}$,
$g_{x\rightarrow r}$ and $g_{i}$ were 128, 128, 128, and 32, respectively.
The network was trained for 800 epochs from scratch using Adam optimizer
\cite{kingma2014adam}. The initial learning rate was $10^{-4}$ with
decay of 0.1 after 250 and 500 epochs. The mini-batch size was 24.
As for our method, 16 images were the first pseudo pairs and 8 images
were the second pseudo pairs. The EMA decay of MT was 0.999. We used
whole images for training rather than cropped patches \cite{xing21invertible}.
In the training, the input images were resized into $256\times256$,
and random flip and rotation were applied. All RAW images were normalized
into {[}0, 1{]} using the black-level and white-level and applied
the bilinear demosaicing for both training and evaluation. 

\subsection{Results}

We compared our method against several state-of-the-art methods: \textbf{UPI}
\cite{brooks2019unprocessing}, a model based invertible ISP. We used
the official parameters determined by metadata of Darmstadt Noise
Dataset \cite{plotz2017benchmarking}. \textbf{CycleISP} \cite{zamir2020cycleisp},
a DNN-based reversed ISP method. We utilized their pre-trained rgb2raw
joint model trained with MIT-Adobe-FiveK Dataset and SIDD. The final
mosacing function was removed for our evaluation. \textbf{U-Net} \cite{conde2022aim},
a simple U-Net \cite{ronneberger2015u} based method used in \cite{conde2022aim}
as a baseline. We removed the final interpolation layer for mosaicing.
\textbf{MBISPLD }\cite{conde2022model}, a hybrid method of the model-based
and learning-based approach. We implemented this method by ourselves
because there was no public code at that time. Our implemented MBISPLD
consisted of unified Gain\&WB, CC, GM with a single parameter, and
TM with static 1x1 convolutions. Note that we did not use the Mosaicing
and Lens-shading blocks as with our method. As for U-Net and MBISPLD,
we trained them from scratch for each dataset using real RGB and RAW
pairs with the same training settings of our method.

We also evaluated several variations of our method. We trained our
model with the real image pairs same as U-Net and MBISPLD instead
of the $\mathrm{PP_{\mathit{rand}}}$. We denote it as ``Real''
and the original setting as ``Pseudo'' training. ``Ours w/o $\mathrm{PP_{\mathit{MT}}}$''
is the model trained without the $\mathrm{PP_{\mathit{MT}}}$. ``Ours
(Flickr/COCO)'' is the model trained with the $\mathrm{PP_{\mathit{MT}}}$
generated from each RAW dataset and Flickr or COCO. ``Ours (All)''
is the model trained with all RAW and RGB datasets.

\subsubsection{RAW Image Reconstruction}

It is difficult to evaluate how well the input RGB image is mapped
to the RAW image similar to the reference RAW image quantitatively
because the perfect ground truth cannot be created in principle. Hence,
in this evaluation, we divided each RAW and RGB image in half, and
used the left RAW image as a reference $y_{ref}$, the right RGB image
as an input $x$, and the right RAW image as a ground truth $y$ with
the assumption that the left and right region has the same characteristics
of the sensing device and lighting. Table \ref{tab:known_ISP} shows
reconstruction results on each dataset in terms of PSNR {[}dB{]} and
the mean Angular Error (AE) {[}$\lyxmathsym{\textdegree}${]} \cite{hordley2004re}
between the predicted color and the ture color. All our method achieved
lower AE, which indicates more accurate reproduction of the global
illumination, than the other methods. It is also shown that our proposed
$\mathrm{PP_{\mathit{MT}}}$ (Flickr/COCO/All) reduced AE when only
pseudo image pairs (Pseudo) were used. This is because the influence
of the difference between the pseudo RGB images generated by $\mathrm{ISP}_{rand}$
and real RGB images was reduced by $\mathrm{PP_{\mathit{MT}}}$. Furthermore,
it is surprising that ours (All) achieved the comparable PSNR and
lower AE compared to the other methods learned with the real pairs.
We consider our method benefited from the data volume of multiple
datasets thanks to the flexible pipeline based on the proposed DPS.
In the setting using the real image pairs (Real), our methods further
outperformed the other methods. It was hard for the other methods
to solve the one-to-many mapping. On the other hand, ours was able
to solve it by reformulating it as one-to-one mapping using the reference
image.


\subsubsection{Robustness to Unknown ISPs \& Sensors}

Table \ref{tab:unknown_ISP1} shows the results of evaluating each
method learned in Table \ref{tab:known_ISP} against MIT-Adobe FiveK
dataset. We chose RGB images generated by unknown ISPs manually tuned
by expert C per image as input RGB images. The proposed method reconstructed
them with higher accuracy than the other methods. The proposed models
learned by the pseudo pairs were more accurate than those learned
by the real pairs. This indicates that the training without the assumption
of a single pipeline is a key to realize the generalized parameter
control. For a similar evaluation, Table \ref{tab:unknown_ISP2} shows
the results of applying the model learned on each dataset to another
dataset. Note that FiveK, SIDD, and LOD used different ISPs for rendering
RGB images. Our method generalized well to arbitrary ISPs and sensors,
although other methods were only accurate for the learned ISP.

\begin{table}
\begin{center}\scalebox{0.8}{

\begin{tabular}{c|l|c|c|c|c}
\hline 
\multirow{2}{*}{Train} & \multirow{2}{*}{Method} & \multicolumn{2}{c|}{Nikon Expert C} & \multicolumn{2}{c}{Canon Expert C}\tabularnewline
\cline{3-6} \cline{4-6} \cline{5-6} \cline{6-6} 
 &  & AE$\downarrow$ & PSNR$\uparrow$ & AE$\downarrow$ & PSNR$\uparrow$\tabularnewline
\hline 
\multicolumn{1}{c}{} & UPI \cite{brooks2019unprocessing} & 10.73  & 26.31  & 11.01  & 27.12 \tabularnewline
\multicolumn{1}{c}{} & CycleISP \cite{zamir2020cycleisp} & 11.01  & 21.73  & 11.31  & 22.11 \tabularnewline
\hline 
 & U-Net \cite{conde2022aim} & 8.28  & 19.81  & 9.08  & 20.35 \tabularnewline
Real & MBISPLD \cite{conde2022model} & 7.61  & 19.36  & 8.64  & 20.26 \tabularnewline
 & Ours w/o $\mathrm{PP_{\mathit{MT}}}$ & 9.62  & 23.47  & 7.15  & 28.65 \tabularnewline
\hline 
\multirow{4}{*}{Pseudo} & Ours w/o $\mathrm{PP_{\mathit{MT}}}$ & 6.22  & 31.20  & 5.39  & 33.35 \tabularnewline
 & Ours (Flickr) & 5.43  & \textbf{32.12 } & 5.06  & 33.61 \tabularnewline
 & Ours (COCO) & 5.36  & 31.77  & 4.99  & 33.22 \tabularnewline
 & Ours (All) & \textbf{4.68}  & 31.53  & \textbf{4.53}  & \textbf{33.67} \tabularnewline
\hline 
\end{tabular}}\end{center}

\caption{Quantitative RAW reconstuction results on the images processed by
the unknown ISP tuned by an expert \cite{fivek}.}

\label{tab:unknown_ISP1}
\end{table}


\begin{table}
\begin{center}\scalebox{0.8}{

\begin{tabular}{cc|l|c|c|c|c}
\hline 
\multirow{2}{*}{Train} & \multicolumn{1}{c|}{} & \multirow{2}{*}{Method} & \multicolumn{4}{c}{Test PSNR$\uparrow$}\tabularnewline
\cline{4-7} \cline{5-7} \cline{6-7} \cline{7-7} 
 &  &  & Nikon & Canon & SIDD & LOD\tabularnewline
\hline 
\multicolumn{1}{c}{} & R & U-Net & 38.01  & \textbf{39.62} & 33.84 & 19.00\tabularnewline
Nikon & R & MBISPLD & \textbf{38.38} & 39.18 & 33.60 & 18.63\tabularnewline
 & P & Ours (Flickr) & 35.51 & 37.77 & \textbf{42.81} & \textbf{34.11}\tabularnewline
\hline 
\multicolumn{1}{c}{} & R & U-Net & 36.93 & \textbf{41.53} & 35.02 & 19.52\tabularnewline
Canon & R & MBISPLD & \textbf{37.39} & 41.35 & 34.77 & 19.27\tabularnewline
 & P & Ours (Flickr) & 35.35 & 37.97 & \textbf{43.19} & \textbf{34.43}\tabularnewline
\hline 
\multicolumn{1}{c}{} & R & U-Net & 29.98 & 32.16 & 45.44 & 21.46\tabularnewline
SIDD & R & MBISPLD & 30.74 & 32.85 & \textbf{45.56} & 21.05\tabularnewline
 & P & Ours (Flickr) & \textbf{34.70} & \textbf{37.49} & 43.68 & \textbf{32.37}\tabularnewline
\hline 
\multicolumn{1}{c}{} & R & U-Net & 24.08 & 26.05 & 35.32 & \textbf{38.94}\tabularnewline
LOD & R & MBISPLD & 24.13 & 26.16 & 35.22 & 37.73\tabularnewline
 & P & Ours (Flickr) & \textbf{37.69} & \textbf{40.18} & \textbf{44.99} & 34.64\tabularnewline
\hline 
\end{tabular}}\end{center}

\caption{Cross dataset evaluation to test genralization to different cameras.
\textquotedblleft R\textquotedblright{} and \textquotedblleft P\textquotedblright{}
denote the real-pair and the $\mathrm{PP}_{rand}$ training, respectively.}

\label{tab:unknown_ISP2}
\end{table}


\subsubsection{Unpaired Evaluation}

We evaluated whether each method was able to convert the actual RGB
datasets (i.e., Flickr and COCO) to RAW-like datasets. Since there
is no ground truth, we compared the distribution of pixel values between
all reference and generated RAW images. Specifically, we used Lab
Histogram Intersections (HI) that are used to compare the color marginal
distributions of images in the Lab color space \cite{isola2017image}.
Table \ref{tab:unpaired_eval} reports the average histogram intersection
over all channels of all pixels. The proposed method using the two
types of pseudo pairs produced the closest distribution to the reference
compared to the other methods. The score of ours (All) was worse on
SIDD compared to ours learned for each dataset. The SIDD differs from
other datasets in that it has fewer images and contains significantly
dark RAW images. Those samples are minor in the mixed dataset, and
it might have caused the degradation of ours (All). However, ours
(All) still achieved better reconstruction results compared to the
other methods.

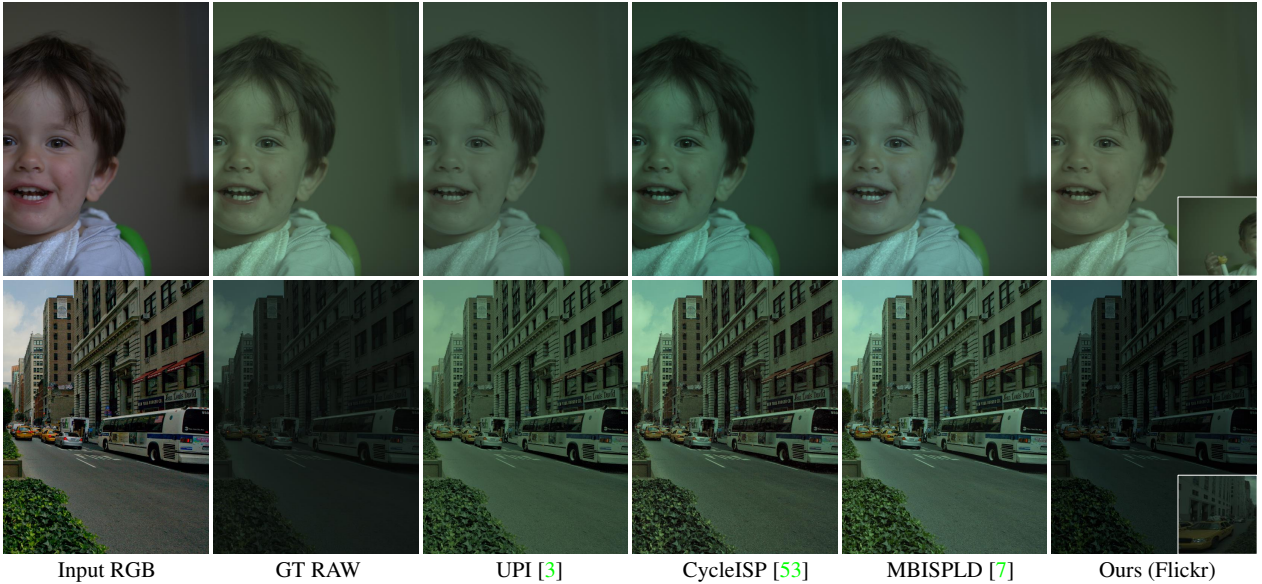
\begin{figure*}
\centering

\def\svgwidth{2.0\columnwidth}

\scriptsize
\begingroup%
  \makeatletter%
  \providecommand\color[2][]{%
    \errmessage{(Inkscape) Color is used for the text in Inkscape, but the package 'color.sty' is not loaded}%
    \renewcommand\color[2][]{}%
  }%
  \providecommand\transparent[1]{%
    \errmessage{(Inkscape) Transparency is used (non-zero) for the text in Inkscape, but the package 'transparent.sty' is not loaded}%
    \renewcommand\transparent[1]{}%
  }%
  \providecommand\rotatebox[2]{#2}%
  \newcommand*\fsize{\dimexpr\f@size pt\relax}%
  \newcommand*\lineheight[1]{\fontsize{\fsize}{#1\fsize}\selectfont}%
  \ifx\svgwidth\undefined%
    \setlength{\unitlength}{954.26495361bp}%
    \ifx\svgscale\undefined%
      \relax%
    \else%
      \setlength{\unitlength}{\unitlength * \real{\svgscale}}%
    \fi%
  \else%
    \setlength{\unitlength}{\svgwidth}%
  \fi%
  \global\let\svgwidth\undefined%
  \global\let\svgscale\undefined%
  \makeatother%
  \begin{picture}(1,0.4620483)%
    \lineheight{1}%
    \setlength\tabcolsep{0pt}%
    \put(0.04305722,0.0036096){\color[rgb]{0,0,0}\makebox(0,0)[lt]{\lineheight{1.25}\smash{\begin{tabular}[t]{l}\scalebox{1.2}{Input RGB}\end{tabular}}}}%
    \put(0.21699424,0.0036096){\color[rgb]{0,0,0}\makebox(0,0)[lt]{\lineheight{1.25}\smash{\begin{tabular}[t]{l}\scalebox{1.2}{GT RAW}\end{tabular}}}}%
    \put(0.70414127,0.0036096){\color[rgb]{0,0,0}\makebox(0,0)[lt]{\lineheight{1.25}\smash{\begin{tabular}[t]{l}\scalebox{1.2}{MBISPLD \cite{conde2022model}}\end{tabular}}}}%
    \put(0.87359383,0.0036096){\color[rgb]{0,0,0}\makebox(0,0)[lt]{\lineheight{1.25}\smash{\begin{tabular}[t]{l}\scalebox{1.2}{Ours (Flickr)}\end{tabular}}}}%
    \put(0.3925821,0.0036096){\color[rgb]{0,0,0}\makebox(0,0)[lt]{\lineheight{1.25}\smash{\begin{tabular}[t]{l}\scalebox{1.2}{UPI \cite{brooks2019unprocessing}}\end{tabular}}}}%
    \put(0.54161827,0.0036096){\color[rgb]{0,0,0}\makebox(0,0)[lt]{\lineheight{1.25}\smash{\begin{tabular}[t]{l}\scalebox{1.2}{CycleISP \cite{zamir2020cycleisp}}\end{tabular}}}}%
    \put(0,0){\includegraphics[width=\unitlength,page=1]{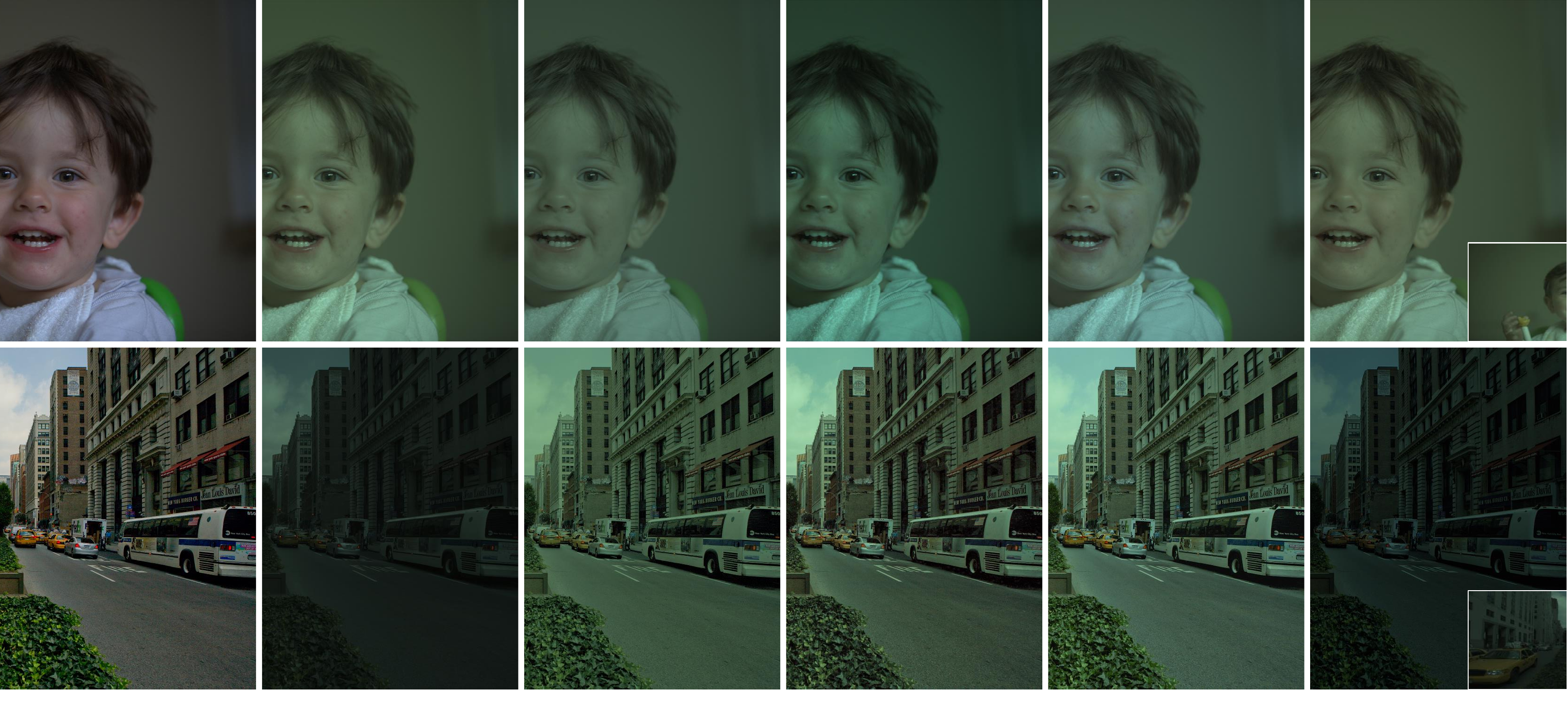}}%
  \end{picture}%
\endgroup%

\caption{Qualitative RAW reconstruction results for Canon EOS 5D. The input
RGB in the first row was processed by Libraw, and that in the second
row was processed by the unknow ISP (expert tuning). The small images
on ours are the reference images.}

\label{fig:result_images}
\end{figure*}


\begin{table*}
\begin{center}\scalebox{0.9}{

\begin{tabular}{c|l|c|c|c|c|c|c|c|c}
\hline 
\multirow{2}{*}{Train} & \multirow{2}{*}{Method} & \multicolumn{4}{c|}{Flickr Histogram Intersection$\uparrow$} & \multicolumn{4}{c}{COCO Histogram Intersection$\uparrow$}\tabularnewline
\cline{3-10} \cline{4-10} \cline{5-10} \cline{6-10} \cline{7-10} \cline{8-10} \cline{9-10} \cline{10-10} 
 &  & Nikon & Canon & SIDD & LOD & Nikon & Canon & SIDD & LOD\tabularnewline
\hline 
\multicolumn{1}{c}{} & UPI \cite{brooks2019unprocessing} & 0.731 & 0.761 & 0.707 & 0.662 & 0.692 & 0.732 & 0.659 & 0.640\tabularnewline
\multicolumn{1}{c}{} & CycleISP \cite{zamir2020cycleisp} & 0.408 & 0.389 & 0.511 & 0.398 & 0.407 & 0.390 & 0.518 & 0.401\tabularnewline
\hline 
\multirow{4}{*}{Real} & U-Net \cite{conde2022aim} & 0.791 & 0.770 & 0.719 & 0.903 & 0.772 & 0.755 & 0.685 & 0.903\tabularnewline
 & MBISPLD \cite{conde2022model} & 0.795 & 0.768 & 0.731 & 0.883 & 0.785 & 0.765 & 0.715 & 0.896\tabularnewline
 & Ours w/o $\mathrm{PP_{\mathit{MT}}}$ & 0.715 & 0.822 & 0.688 & 0.899 & 0.686 & 0.817 & 0.651 & 0.899\tabularnewline
 & Ours (Flickr/COCO) & 0.851 & 0.895 & 0.762 & \textbf{0.965} & 0.877 & 0.921 & 0.799 & \textbf{0.967}\tabularnewline
\hline 
\multirow{3}{*}{Pseudo} & Ours w/o $\mathrm{PP_{\mathit{MT}}}$ & 0.909 & 0.900 & 0.849 & 0.897 & 0.905 & 0.924 & 0.841 & 0.897\tabularnewline
 & Ours (Flickr/COCO) & 0.930 & 0.940 & \textbf{0.852} & 0.959 & 0.931 & \textbf{0.946} & \textbf{0.845} & 0.959\tabularnewline
 & Ours (All) & \textbf{0.937} & \textbf{0.945} & 0.816 & 0.949 & \textbf{0.935} & \textbf{0.946} & 0.812 & 0.952\tabularnewline
\hline 
\end{tabular}}\end{center}

\caption{Histogram Intersection (HI) in Lab color space between the generated
RAW images and the reference RAW images.}

\label{tab:unpaired_eval}
\end{table*}


\subsubsection{Ablation Studies}

Table \ref{tab:ablation} shows the effectiveness of our proposed
modules, i.e., DPS, GG, DTM, and $\mathrm{PP}_{\mathit{MT}}$, evaluated
on the Canon EOS 5D. Each module contributed to the performance. In
particular, the effect of DPS was significant (-5.87$\lyxmathsym{\textdegree}$,
+8.27dB), followed by DTM (-0.55$\lyxmathsym{\textdegree}$, +1.02dB)
and GG (-0.04$\lyxmathsym{\textdegree}$, +1.14dB). The $\mathrm{PP}_{\mathit{MT}}$
mainly contributed to generalization to unknown ISPs. On the other
hand, all metrics were degraded if we replaced $\mathrm{PP}_{\mathit{MT}}$
with $\mathrm{PP_{\mathit{MT}}^{-}}$ or SL. From the result, we concluded
that the proposed $\mathrm{PP}_{\mathit{MT}}$ effectively realized
learning on the unpaired data.

\begin{table}
\begin{center}\scalebox{0.9}{

\begin{tabular}{cccl|c|c|c}
\hline 
\multicolumn{4}{c|}{Module} & \multicolumn{2}{c|}{Canon EOS 5D} & Flickr\tabularnewline
\hline 
DPS & GG & DTM & UT & AE$\downarrow$ & PSNR$\uparrow$ & HI$\uparrow$\tabularnewline
\hline 
 &  &  &  & 10.46 & 27.47 & 0.815\tabularnewline
$\checkmark$ &  &  &  & 4.59 & 35.74 & 0.898\tabularnewline
$\checkmark$ & $\checkmark$ &  &  & 4.55 & 36.87 & 0.901\tabularnewline
$\checkmark$ & $\checkmark$ & $\checkmark$ &  & 4.00 & 37.89 & 0.900\tabularnewline
$\checkmark$ & $\checkmark$ & $\checkmark$ & SL & 4.17 & 37.16 & 0.908\tabularnewline
$\checkmark$ & $\checkmark$ & $\checkmark$ & $\mathrm{PP_{\mathit{MT}}^{-}}$ & 4.97 & 35.64 & 0.907\tabularnewline
$\checkmark$ & $\checkmark$ & $\checkmark$ & $\mathrm{PP_{\mathit{MT}}}$ & \textbf{3.72} & \textbf{37.97} & \textbf{0.940}\tabularnewline
\hline 
\end{tabular}}\end{center}

\caption{Ablation studies of the proposed modules. UT denotes Unpaired Training.
$\mathrm{PP}_{\mathit{MT}}$ is the proposed second pseudo-pair training,
and $\mathrm{PP_{\mathit{MT}}^{-}}$ is the version that uses the
target RAW image as the reference image for the student. SL is Style
Loss \cite{johnson2016perceptual}}

\label{tab:ablation}
\end{table}


\subsubsection{Quanlitative Results}

We show qualitative comparisons against the other methods in Figure
\ref{fig:result_images}. The first and the second row show the results
for the normal Canon EOS 5D and the Expert C image, respectively.
The small images on the results of our method are the given reference
images. The proposed method produced RAW-like images that have less
color or brightness misalignment than the other methods. Although
MBISPLD learned with the real pairs of Canon EOS 5D, it failed to
reproduce the ambient light color that is removed in the input RGB
image. The Expert C image was manually adjusted to be bright, so the
other methods resulted in producing brighter images than the real
RAW image. On the other hand, our method produced more GT-like RAW
image thanks to the reference guidance.

\subsubsection{RAW Image Object Detection}

We used the LOD RAW-like images converted from the COCO's RGB images
by each method to learn 8 class object detection on the LOD dataset.
Following \cite{Hong2021Crafting}, CenterNet \cite{zhou2019objects}
model pre-trained with COCO's RGB images was fine-tuned using the
LOD RAW-like images. The details are in the supplementary material.
Table \ref{tab:det_LOD} shows the detection accuracy (mAP) of the
trained model on the real LOD RAW images for normal-light condition.
MBISPLD with the dictionary augmentation \cite{conde2022model} (+DA)
was also evaluated. Ours achieved the greatest accuracy improvement
although all methods improved the pre-trained model. Our method successfully
generated pseudo-RAW images effective for training in the down-stream
task without the paired RGB and RAW images or metadata.

\begin{table}
\begin{center}\scalebox{0.9}{

\begin{tabular}{l|c}
\hline 
\multirow{1}{*}{Fine-tuning data} & \multicolumn{1}{c}{mAP@0.5:0.95}\tabularnewline
\hline 
No fine-tuning & 41.60\tabularnewline
UPI \cite{conde2022aim} & 46.03\tabularnewline
MBISPLD \cite{conde2022model} & 46.37\tabularnewline
MBISPLD+DA \cite{conde2022model} & 46.47\tabularnewline
Ours (COCO) & \textbf{47.17}\tabularnewline
\hline 
\end{tabular}}\end{center}

\caption{Object detection accuracy on LOD Dataset with fine-tuning using RAW-like
data converted from COCO Dataset}

\label{tab:det_LOD}
\end{table}


\section{Conclusion}

In this paper, we have proposed a self-supervised reversed ISP method
that does not require metadata and paired images. The proposed method
is able to handle diverse RGB-to-RAW mappings by learning how to control
the mapping parameters based on a given reference RAW image. Furthermore,
the entire pipeline is trainable using unpaired RGB and RAW images.
The experiments showed the proposed method successfully produced the
target RAW-like images. We hope this approach will contribute to the
future progress of the RAW image-based computer vision research.


{\small{}\bibliographystyle{ieee_fullname}
\bibliography{egbib}

\begin{thebibliography}{10}\itemsep=-1pt

\bibitem{sidd}
Abdelrahman Abdelhamed, Stephen Lin, and Michael~S. Brown.
\newblock A high-quality denoising dataset for smartphone cameras.
\newblock In {\em 2018 IEEE/CVF Conference on Computer Vision and Pattern
  Recognition}, pages 1692--1700, 2018.

\bibitem{afifi2021cie}
Mahmoud Afifi, Abdelrahman Abdelhamed, Abdullah Abuolaim, Abhijith
  Punnappurath, and Michael~S Brown.
\newblock Cie xyz net: Unprocessing images for low-level computer vision tasks.
\newblock {\em IEEE Transactions on Pattern Analysis and Machine Intelligence},
  44(9):4688--4700, 2021.

\bibitem{brooks2019unprocessing}
Tim Brooks, Ben Mildenhall, Tianfan Xue, Jiawen Chen, Dillon Sharlet, and
  Jonathan~T Barron.
\newblock Unprocessing images for learned raw denoising.
\newblock In {\em Proceedings of the IEEE/CVF Conference on Computer Vision and
  Pattern Recognition}, pages 11036--11045, 2019.

\bibitem{buckler2017reconfiguring}
Mark Buckler, Suren Jayasuriya, and Adrian Sampson.
\newblock Reconfiguring the imaging pipeline for computer vision.
\newblock In {\em Proceedings of the IEEE International Conference on Computer
  Vision}, pages 975--984, 2017.

\bibitem{fivek}
Vladimir Bychkovsky, Sylvain Paris, Eric Chan, and Fr{\'e}do Durand.
\newblock Learning photographic global tonal adjustment with a database of
  input / output image pairs.
\newblock In {\em The Twenty-Fourth IEEE Conference on Computer Vision and
  Pattern Recognition}, 2011.

\bibitem{chen2020dynamicc}
Yinpeng Chen, Xiyang Dai, Mengchen Liu, Dongdong Chen, Lu Yuan, and Zicheng
  Liu.
\newblock Dynamic convolution: Attention over convolution kernels.
\newblock In {\em Proceedings of the IEEE/CVF Conference on Computer Vision and
  Pattern Recognition}, pages 11030--11039, 2020.

\bibitem{conde2022model}
Marcos~V Conde, Steven McDonagh, Matteo Maggioni, Ales Leonardis, and Eduardo
  P{\'e}rez-Pellitero.
\newblock Model-based image signal processors via learnable dictionaries.
\newblock In {\em Proceedings of the AAAI Conference on Artificial
  Intelligence}, volume~36, pages 481--489, 2022.

\bibitem{conde2022aim}
Marcos~V Conde, Radu Timofte, et~al.
\newblock {R}eversed {I}mage {S}ignal {P}rocessing and {RAW} {R}econstruction.
  {AIM} 2022 {C}hallenge {R}eport.
\newblock In {\em Proceedings of the European Conference on Computer Vision
  Workshops (ECCVW)}, 2022.

\bibitem{Cui_2022_BMVC}
Ziteng Cui, Kunchang Li, Lin Gu, Shenghan Su, Peng Gao, ZhengKai Jiang, Yu
  Qiao, and Tatsuya Harada.
\newblock You only need 90k parameters to adapt light: a light weight
  transformer for image enhancement and exposure correction.
\newblock In {\em 33rd British Machine Vision Conference 2022, {BMVC} 2022,
  London, UK, November 21-24, 2022}. {BMVA} Press, 2022.

\bibitem{dong2023rispnet}
Xiaoyi Dong, Yu Zhu, Chenghua Li, Peisong Wang, and Jian Cheng.
\newblock Rispnet: a network for reversed image signal processing.
\newblock In {\em Computer Vision--ECCV 2022 Workshops: Tel Aviv, Israel,
  October 23--27, 2022, Proceedings, Part II}, pages 445--457. Springer, 2023.

\bibitem{ebner2007color}
Marc Ebner.
\newblock {\em Color constancy}, volume~7.
\newblock John Wiley \& Sons, 2007.

\bibitem{hansen1996adapting}
Nikolaus Hansen and Andreas Ostermeier.
\newblock Adapting arbitrary normal mutation distributions in evolution
  strategies: The covariance matrix adaptation.
\newblock In {\em Proceedings of IEEE international conference on evolutionary
  computation}, pages 312--317. IEEE, 1996.

\bibitem{hevia2020optimization}
Luis~V Hevia, Miguel~A Patricio, Jose{\'e}~M Molina, and Antonio Berlanga.
\newblock Optimization of the isp parameters of a camera through differential
  evolution.
\newblock {\em IEEE Access}, 8:143479--143493, 2020.

\bibitem{ho2019flow++}
Jonathan Ho, Xi Chen, Aravind Srinivas, Yan Duan, and Pieter Abbeel.
\newblock Flow++: Improving flow-based generative models with variational
  dequantization and architecture design.
\newblock In {\em International Conference on Machine Learning}, pages
  2722--2730. PMLR, 2019.

\bibitem{Hong2021Crafting}
Yang Hong, Kaixuan Wei, Linwei Chen, and Ying Fu.
\newblock Crafting object detection in very low light.
\newblock In {\em Proceedings of the British Machine Vision Virtual
  Conference}, 2021.

\bibitem{hordley2004re}
Steven~D Hordley and Graham~D Finlayson.
\newblock Re-evaluating colour constancy algorithms.
\newblock In {\em Proceedings of the 17th International Conference on Pattern
  Recognition, 2004. ICPR 2004.}, volume~1, pages 76--79. IEEE, 2004.

\bibitem{ignatov2020replacing}
Andrey Ignatov, Luc Van~Gool, and Radu Timofte.
\newblock Replacing mobile camera isp with a single deep learning model.
\newblock In {\em Proceedings of the IEEE/CVF Conference on Computer Vision and
  Pattern Recognition Workshops}, pages 536--537, 2020.

\bibitem{isola2017image}
Phillip Isola, Jun-Yan Zhu, Tinghui Zhou, and Alexei~A Efros.
\newblock Image-to-image translation with conditional adversarial networks.
\newblock In {\em Proceedings of the IEEE conference on computer vision and
  pattern recognition}, pages 1125--1134, 2017.

\bibitem{johnson2016perceptual}
Justin Johnson, Alexandre Alahi, and Li Fei-Fei.
\newblock Perceptual losses for real-time style transfer and super-resolution.
\newblock In {\em Computer Vision--ECCV 2016: 14th European Conference,
  Amsterdam, The Netherlands, October 11-14, 2016, Proceedings, Part II 14},
  pages 694--711. Springer, 2016.

\bibitem{karaimer2016software}
Hakki~Can Karaimer and Michael~S Brown.
\newblock A software platform for manipulating the camera imaging pipeline.
\newblock In {\em Computer Vision--ECCV 2016: 14th European Conference,
  Amsterdam, The Netherlands, October 11--14, 2016, Proceedings, Part I 14},
  pages 429--444. Springer, 2016.

\bibitem{kim2023overexposure}
Jinha Kim, Jun Jiang, and Jinwei Gu.
\newblock Overexposure mask fusion: Generalizable reverse isp multi-step
  refinement.
\newblock In {\em Computer Vision--ECCV 2022 Workshops: Tel Aviv, Israel,
  October 23--27, 2022, Proceedings, Part II}, pages 699--713. Springer, 2023.

\bibitem{kim2012new}
Seon~Joo Kim, Hai~Ting Lin, Zheng Lu, Sabine S{\"u}sstrunk, Stephen Lin, and
  Michael~S Brown.
\newblock A new in-camera imaging model for color computer vision and its
  application.
\newblock {\em IEEE Transactions on Pattern Analysis and Machine Intelligence},
  34(12):2289--2302, 2012.

\bibitem{kingma2014adam}
Diederik~P Kingma and Jimmy Ba.
\newblock Adam: A method for stochastic optimization.
\newblock {\em arXiv preprint arXiv:1412.6980}, 2014.

\bibitem{kingma2018glow}
Durk~P Kingma and Prafulla Dhariwal.
\newblock Glow: Generative flow with invertible 1x1 convolutions.
\newblock {\em Advances in neural information processing systems}, 31, 2018.

\bibitem{kinli2023reversing}
Furkan K{\i}nl{\i}, Bar{\i}{\c{s}} {\"O}zcan, and Furkan K{\i}ra{\c{c}}.
\newblock Reversing image signal processors by reverse style transferring.
\newblock In {\em Computer Vision--ECCV 2022 Workshops: Tel Aviv, Israel,
  October 23--27, 2022, Proceedings, Part II}, pages 688--698. Springer, 2023.

\bibitem{li2021synthetic}
Chen Li and Gim~Hee Lee.
\newblock From synthetic to real: Unsupervised domain adaptation for animal
  pose estimation.
\newblock In {\em Proceedings of the IEEE/CVF conference on computer vision and
  pattern recognition}, pages 1482--1491, 2021.

\bibitem{li2022cross}
Yu-Jhe Li, Xiaoliang Dai, Chih-Yao Ma, Yen-Cheng Liu, Kan Chen, Bichen Wu,
  Zijian He, Kris Kitani, and Peter Vajda.
\newblock Cross-domain adaptive teacher for object detection.
\newblock In {\em IEEE Conference on Computer Vision and Pattern Recognition},
  2022.

\bibitem{liang2020raw}
Chih-Hung Liang, Yu-An Chen, Yueh-Cheng Liu, and Winston~H Hsu.
\newblock Raw image deblurring.
\newblock {\em IEEE Transactions on Multimedia}, 24:61--72, 2020.

\bibitem{lin2014microsoft}
Tsung-Yi Lin, Michael Maire, Serge Belongie, James Hays, Pietro Perona, Deva
  Ramanan, Piotr Doll{\'a}r, and C~Lawrence Zitnick.
\newblock Microsoft coco: Common objects in context.
\newblock In {\em European conference on computer vision}, pages 740--755.
  Springer, 2014.

\bibitem{liu2018darts}
Hanxiao Liu, Karen Simonyan, and Yiming Yang.
\newblock Darts: Differentiable architecture search.
\newblock {\em arXiv preprint arXiv:1806.09055}, 2018.

\bibitem{liu2022deep}
Shuai Liu, Chaoyu Feng, Xiaotao Wang, Hao Wang, Ran Zhu, Yongqiang Li, and Lei
  Lei.
\newblock Deep-flexisp: A three-stage framework for night photography
  rendering.
\newblock In {\em Proceedings of the IEEE/CVF Conference on Computer Vision and
  Pattern Recognition}, pages 1211--1220, 2022.

\bibitem{liu2021unbiased}
Yen-Cheng Liu, Chih-Yao Ma, Zijian He, Chia-Wen Kuo, Kan Chen, Peizhao Zhang,
  Bichen Wu, Zsolt Kira, and Peter Vajda.
\newblock Unbiased teacher for semi-supervised object detection.
\newblock In {\em Proceedings of the International Conference on Learning
  Representations}, 2021.

\bibitem{Liu_2022_CVPR}
Yen-Cheng Liu, Chih-Yao Ma, and Zsolt Kira.
\newblock Unbiased teacher v2: Semi-supervised object detection for anchor-free
  and anchor-based detectors.
\newblock In {\em Proceedings of the IEEE/CVF Conference on Computer Vision and
  Pattern Recognition}, pages 9819--9828, June 2022.

\bibitem{loshchilov2016sgdr}
Ilya Loshchilov and Frank Hutter.
\newblock Sgdr: Stochastic gradient descent with warm restarts.
\newblock {\em arXiv preprint arXiv:1608.03983}, 2016.

\bibitem{huiskes08}
B.~Thomee Mark J.~Huiskes and Michael~S. Lew.
\newblock New trends and ideas in visual concept detection: The mir flickr
  retrieval evaluation initiative.
\newblock In {\em MIR '10: Proceedings of the 2010 ACM International Conference
  on Multimedia Information Retrieval}, pages 527--536, New York, NY, USA,
  2010. ACM.

\bibitem{morawski2022genisp}
Igor Morawski, Yu-An Chen, Yu-Sheng Lin, Shusil Dangi, Kai He, and Winston~H
  Hsu.
\newblock Genisp: Neural isp for low-light machine cognition.
\newblock In {\em Proceedings of the IEEE/CVF Conference on Computer Vision and
  Pattern Recognition}, pages 630--639, 2022.

\bibitem{mosleh2020hardware}
Ali Mosleh, Avinash Sharma, Emmanuel Onzon, Fahim Mannan, Nicolas Robidoux, and
  Felix Heide.
\newblock Hardware-in-the-loop end-to-end optimization of camera image
  processing pipelines.
\newblock In {\em Proceedings of the IEEE/CVF Conference on Computer Vision and
  Pattern Recognition}, pages 7529--7538, 2020.

\bibitem{nishimura2018automatic}
Jun Nishimura, Timo Gerasimow, Rao Sushma, Aleksandar Sutic, Chyuan-Tyng Wu,
  and Gilad Michael.
\newblock Automatic isp image quality tuning using nonlinear optimization.
\newblock In {\em 2018 25th IEEE International Conference on Image Processing
  (ICIP)}, pages 2471--2475. IEEE, 2018.

\bibitem{onzon2021neural}
Emmanuel Onzon, Fahim Mannan, and Felix Heide.
\newblock Neural auto-exposure for high-dynamic range object detection.
\newblock In {\em Proceedings of the IEEE/CVF Conference on Computer Vision and
  Pattern Recognition}, pages 7710--7720, 2021.

\bibitem{pavithra2021automatic}
G Pavithra and Bhat Radhesh.
\newblock Automatic image quality tuning framework for optimization of isp
  parameters based on multi-stage optimization approach.
\newblock {\em Electronic Imaging}, 2021(9):197--1, 2021.

\bibitem{plotz2017benchmarking}
Tobias Plotz and Stefan Roth.
\newblock Benchmarking denoising algorithms with real photographs.
\newblock In {\em Proceedings of the IEEE conference on computer vision and
  pattern recognition}, pages 1586--1595, 2017.

\bibitem{robidoux2021end}
Nicolas Robidoux, Luis E~Garcia Capel, Dong-eun Seo, Avinash Sharma, Federico
  Ariza, and Felix Heide.
\newblock End-to-end high dynamic range camera pipeline optimization.
\newblock In {\em Proceedings of the IEEE/CVF Conference on Computer Vision and
  Pattern Recognition}, pages 6297--6307, 2021.

\bibitem{ronneberger2015u}
Olaf Ronneberger, Philipp Fischer, and Thomas Brox.
\newblock U-net: Convolutional networks for biomedical image segmentation.
\newblock In {\em Medical Image Computing and Computer-Assisted
  Intervention--MICCAI 2015: 18th International Conference, Munich, Germany,
  October 5-9, 2015, Proceedings, Part III 18}, pages 234--241. Springer, 2015.

\bibitem{schwartz2018deepisp}
Eli Schwartz, Raja Giryes, and Alex~M Bronstein.
\newblock Deepisp: Toward learning an end-to-end image processing pipeline.
\newblock {\em IEEE Transactions on Image Processing}, 28(2):912--923, 2018.

\bibitem{tarvainen2017mean}
Antti Tarvainen and Harri Valpola.
\newblock Mean teachers are better role models: Weight-averaged consistency
  targets improve semi-supervised deep learning results.
\newblock {\em Advances in neural information processing systems}, 30, 2017.

\bibitem{tseng2019hyperparameter}
Ethan Tseng, Felix Yu, Yuting Yang, Fahim Mannan, Karl~ST Arnaud, Derek
  Nowrouzezahrai, Jean-Fran{\c{c}}ois Lalonde, and Felix Heide.
\newblock Hyperparameter optimization in black-box image processing using
  differentiable proxies.
\newblock {\em ACM Trans. Graph.}, 38(4):27--1, 2019.

\bibitem{xing21invertible}
Yazhou Xing, Zian Qian, and Qifeng Chen.
\newblock Invertible image signal processing.
\newblock In {\em CVPR}, 2021.

\bibitem{yang2019condconv}
Brandon Yang, Gabriel Bender, Quoc~V Le, and Jiquan Ngiam.
\newblock Condconv: Conditionally parameterized convolutions for efficient
  inference.
\newblock {\em Advances in Neural Information Processing Systems}, 32, 2019.

\bibitem{yoshimura2022dynamicisp}
Masakazu Yoshimura, Junji Otsuka, Atsushi Irie, and Takeshi Ohashi.
\newblock Dynamicisp: Dynamically controlled image signal processor for image
  recognition.
\newblock {\em arXiv preprint arXiv:2211.01146}, 2022.

\bibitem{yoshimura2022rawgment}
Masakazu Yoshimura, Junji Otsuka, Atsushi Irie, and Takeshi Ohashi.
\newblock Rawgment: Noise-accounted raw augmentation enables recognition in a
  wide variety of environments.
\newblock {\em arXiv preprint arXiv:2210.16046}, 2022.

\bibitem{yu2018deep}
Fisher Yu, Dequan Wang, Evan Shelhamer, and Trevor Darrell.
\newblock Deep layer aggregation.
\newblock In {\em Proceedings of the IEEE conference on computer vision and
  pattern recognition}, pages 2403--2412, 2018.

\bibitem{yu2021reconfigisp}
Ke Yu, Zexian Li, Yue Peng, Chen~Change Loy, and Jinwei Gu.
\newblock Reconfigisp: Reconfigurable camera image processing pipeline.
\newblock In {\em Proceedings of the IEEE/CVF International Conference on
  Computer Vision}, pages 4248--4257, 2021.

\bibitem{zamir2020cycleisp}
Syed~Waqas Zamir, Aditya Arora, Salman Khan, Munawar Hayat, Fahad~Shahbaz Khan,
  Ming-Hsuan Yang, and Ling Shao.
\newblock Cycleisp: Real image restoration via improved data synthesis.
\newblock In {\em Proceedings of the IEEE/CVF Conference on Computer Vision and
  Pattern Recognition}, pages 2696--2705, 2020.

\bibitem{zhang2019zoom}
Xuaner Zhang, Qifeng Chen, Ren Ng, and Vladlen Koltun.
\newblock Zoom to learn, learn to zoom.
\newblock In {\em Proceedings of the IEEE Conference on Computer Vision and
  Pattern Recognition}, 2019.

\bibitem{zhou2019objects}
Xingyi Zhou, Dequan Wang, and Philipp Kr{\"a}henb{\"u}hl.
\newblock Objects as points.
\newblock {\em arXiv preprint arXiv:1904.07850}, 2019.

\bibitem{zou2023learned}
Beiji Zou and Yue Zhang.
\newblock Learned reverse isp with soft supervision.
\newblock In {\em Computer Vision--ECCV 2022 Workshops: Tel Aviv, Israel,
  October 23--27, 2022, Proceedings, Part II}, pages 489--506. Springer, 2023.

\end{thebibliography}
 }{\small\par}

\appendix
\clearpage{}

\part*{Appendices}

\section{Implementation Details}

\subsection{ISP Blocks}

The proposed reversed ISP pipeline consists of five ISP blocks: Global
Gain (GG) $f_{gg}$, White Balance (WB) $f_{wb}$, Color Correction
(CC) $f_{cc}$, Gamma Correction (GC) $f_{gc}$, and Tone Mapping
(TM) $f_{tm}$. In this section, we describe the detailed implementation
of the blocks except for TM.

\subsubsection{Global Gain}

The forward function $f_{\ensuremath{gg}}$ simply multiplies single
scalar $g$ over all pixels to adjust global brightness:

\begin{equation}
f_{gg}(x,g)=gx.
\end{equation}
In MBISPLD \cite{conde2022model}, GG is unified into WB with a single
parameter dictionary. We designed them separately with independent
dictionaries to enhance their representation capability. In addition,
we used a highlight preservation technique \cite{brooks2019unprocessing}. 

\subsubsection{White Balance}

The white balancing is an operation to remove global illumination
effect and achieve color constancy. The $f_{\ensuremath{wb}}$ is
expressed as matrix multiplication of a matrix style input image $\text{X\ensuremath{\in}\ensuremath{\mathbb{R^{\mathrm{HW\times3}}}}}$
and a $3\times3$ diagonal matrix composed of channel-wise gains $[g_{r},g_{g},g_{b}]$:

\begin{equation}
f_{wb}(x,[g_{r},g_{g},g_{b}])=Xdiag([g_{r},g_{g},g_{b}]).
\end{equation}
We initialized $K$ candidates of $g_{r}$ and $g_{b}$ in the parameter
dictionary using a uniform distribution $\mathcal{U}(1,2)$ and set
$g_{g}$ as 1.0 based on the traditional ISP pipeline. The highlight
preservation technique was used as well as GG.

\subsubsection{Color Correction}

The color correction connects the camera specific RAW color space
to the common linear sRGB space. The $f_{\ensuremath{cc}}$ is expressed
as matrix multiplication of $X$ and a $3\times3$ Color Correction
Matrix (CCM) $M_{ccm}$ such that:

\begin{equation}
f_{cc}(x,M_{ccm})=XM_{ccm}.
\end{equation}
To maintain realistic $M_{ccm}$ features, we applied column-normalization
\cite{brooks2019unprocessing,conde2022model} to $M_{ccm}$ at every
inference. 

\subsubsection{Gamma Correction}

Human perception is more sensitive to darker region than brighter
region. To adjust images for human perception, simple gamma correction
is widely used:

\begin{equation}
f_{gc}(x,\gamma)=\max(x,10^{-8})^{1/\gamma}.
\end{equation}
We initialized $K$ candidates of $\gamma$ in the parameter dictionary
using a uniform distribution $\mathcal{U}(1.7,2.7)$ whose center
is 2.2, which is commonly chosen for traditional ISP. 

\subsection{Randomized ISP}

\begin{figure*}
\centering

\def\svgwidth{1.7\columnwidth}

\scriptsize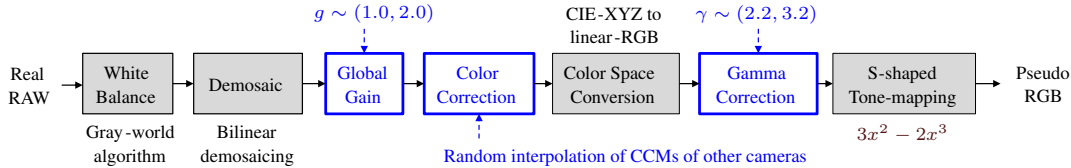

\caption{The pipeline of the randomized traditional ISP ($\mathrm{ISP}_{\mathrm{\mathit{rand}}}$)
for the first pseudo pair ($\mathrm{PP}_{rand}$) training.}

\label{fig:RandISP-1}
\end{figure*}

\begin{figure*}
\centering

\def\svgwidth{2.0\columnwidth}

\scriptsize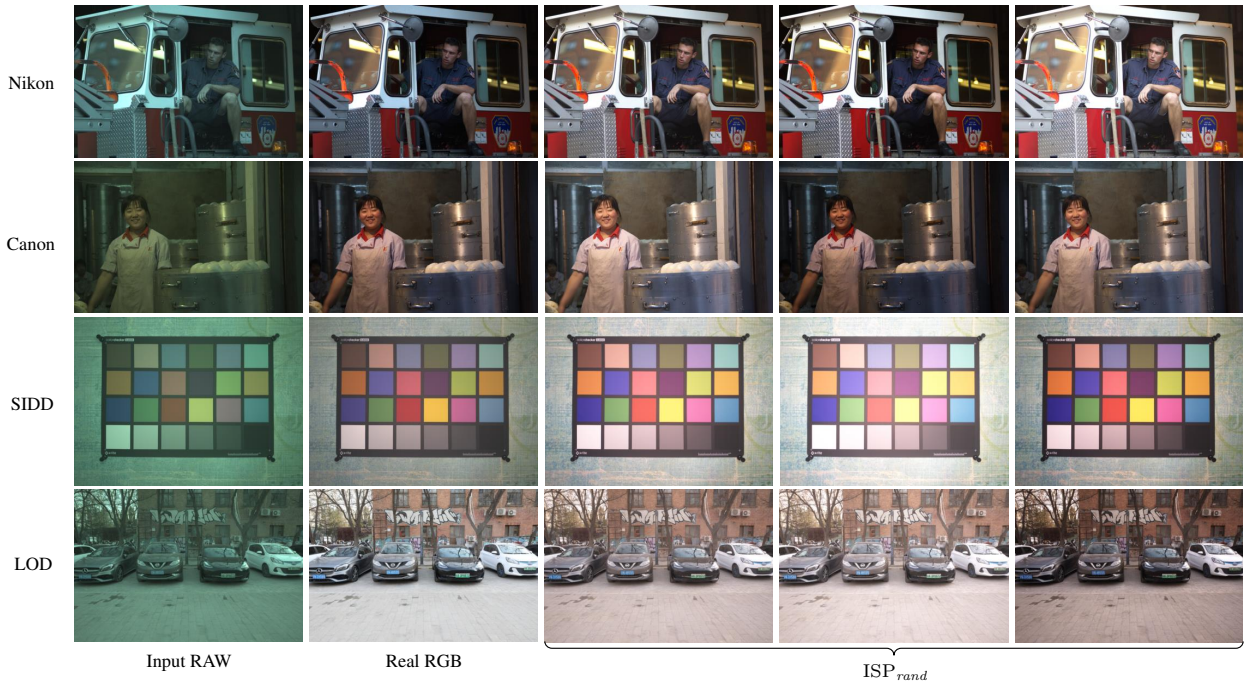

\caption{Pseudo RGB images generated by the randomized traditional ISP ($\mathrm{ISP}_{\mathrm{\mathit{rand}}}$)
for the first pseudo pair ($\mathrm{PP}_{rand}$) training.}

\label{fig:RandISP_examples}
\end{figure*}

Figure \ref{fig:RandISP-1} shows the pipeline of the $\mathrm{ISP}_{\mathrm{\mathit{rand}}}$
that generates the $\mathrm{PP}_{rand}$. We essentially followed
the implementation of UPI \cite{brooks2019unprocessing}, but the
parameters of WB, GG, CC, and GM were determined independently of
the target pipeline. The WB gains were determined by a simple WB algorithm
based on the Gray-world assumption \cite{ebner2007color} without
using any metadata. Specifically, per-channel averages of pixel values
were used to determine the gains as follows:

\begin{equation}
\left(g_{r},g_{g},g_{b}\right)=\left(\frac{G_{mean}}{R_{mean}},1,\frac{G_{mean}}{B_{mean}}\right),
\end{equation}
where $R_{mean}$, $G_{mean}$, and $B_{mean}$ are the average pixel
values for R, G, and B channels, respectively. To remove outliers,
we calculated the luma Y in YCbCr color space first and excluded pixels
whose Y values were top and bottom 5\% for the mean calculating. The
CCM of CC was generated with random interpolation of CCMs of irrelevant
cameras obtained from Darmstadt Noise Dataset \cite{plotz2017benchmarking}
and MIT-Adobe-FiveK Dataset \cite{fivek}. The global gain and gamma
parameters were sampled from uniform distributions $\mathcal{U}(1,2)$
and $\mathcal{U}(2.2,3.2)$, respectively. These values were determined
to make images brighter because the UPI pipleline tends to generate
darker images than the images in COCO \cite{lin2014microsoft} or
Flickr 1 Million Dataset \cite{huiskes08}. Instead of generating
pseudo-RGB images on the fly at training time, we generated one RGB
image from each RGB image before training to speed up experiments.
Figure \ref{fig:RandISP_examples} shows examples of images generated
by the $\mathrm{ISP}_{\mathrm{\mathit{rand}}}$ for the Nikon, Canon,
SIDD, and LOD images. Note that we applied the gamma correction with
$\gamma=2.2$ to all RAW images only for visualization.

\subsection{Evaluation Metrics}

\begin{table*}
\begin{center}\scalebox{0.9}{

\begin{tabular}{c|l|c|c|c|c|c|c|c|c}
\hline 
\multirow{2}{*}{Data} & \multirow{2}{*}{Method} & \multicolumn{2}{c|}{Nikon D700} & \multicolumn{2}{c|}{Canon EOS 5D} & \multicolumn{2}{c|}{SIDD} & \multicolumn{2}{c}{LOD}\tabularnewline
\cline{3-10} \cline{4-10} \cline{5-10} \cline{6-10} \cline{7-10} \cline{8-10} \cline{9-10} \cline{10-10} 
 &  & AE$\downarrow$ & PSNR$\uparrow$ & AE$\downarrow$ & PSNR$\uparrow$ & AE$\downarrow$ & PSNR$\uparrow$ & AE$\downarrow$ & PSNR$\uparrow$\tabularnewline
\hline 
\multirow{7}{*}{Pseudo} & U-Net \cite{conde2022aim} & 9.07 & 24.51 & 9.19 & 26.22 & 9.31 & 35.43 & 7.84 & 31.52\tabularnewline
 & U-Net (All) \cite{conde2022aim} & 7.34 & 24.34 & 6.83 & 26.71 & 10.08 & 36.64 & 8.29 & 31.76\tabularnewline
 & MBISPLD \cite{conde2022model} & 7.35 & 23.66 & 7.32 & 25.63 & 8.53 & 36.95 & 8.84 & 34.10\tabularnewline
 & MBISPLD (All) \cite{conde2022model} & 9.70 & 23.37 & 8.62 & 25.39 & 10.81 & 36.01 & 7.96 & 34.50\tabularnewline
 & Ours (Flickr) & 3.81 & 35.51 & 3.72 & 37.97 & 5.84 & 43.68 & 4.91 & 34.64\tabularnewline
 & Ours (COCO) & 3.61 & 35.52 & 3.59 & 38.36 & 6.00 & 43.51 & \textbf{4.90} & \textbf{34.96}\tabularnewline
 & Ours (All) & \textbf{3.02} & \textbf{38.80} & \textbf{3.20} & \textbf{41.18} & \textbf{5.23} & \textbf{46.23} & 4.91 & 34.89\tabularnewline
\hline 
\end{tabular}}\end{center}

\caption{Quantitative RAW reconstuction results among our methods and other
methods trained without the real paired images. The characters (Flickr/COCO)
in parentheses of ours denote the dataset used for the $\mathrm{PP_{\mathit{MT}}}$.
Ours (All) was trained with the $\mathrm{PP_{\mathit{rand}}}$ and
$\mathrm{PP_{\mathit{MT}}}$ generated from all datasets. U-Net (All)
and MBISPLD (All) were trained with $\mathrm{PP_{\mathit{rand}}}$
generated from all datasets.}

\label{tab:others_with_PP}
\end{table*}

In the evaluation with ground-truth RAW images, PSNR {[}dB{]} and
Angular Error (AE) {[}$\lyxmathsym{\textdegree}${]} were evaluated
with the half split images as we described. That is, we used the left
RAW image as the reference and the right RGB and RAW images as the
input and ground-truth images. When we split the image horizontally,
all images were rotated so that the long side became the width. In
addition, the right RGB and RAW images were further split in half
vertically, and we evaluated them independently with the same reference
image (the left image) to reduce GPU memory usage. This kind of split
evaluation was also employed in \cite{xing21invertible,conde2022model}.
Note that other methods were also evaluated with the same split images.

In the evaluation without ground-truth RAW images, Histogram Intersenction
(HI) in Lab color space was used as an evaluation metric. To compute
HI, first, each RAW pixel value was converted to CIE XYZ color space
using the CCM of the reference RAW image. Second, the CIE XYZ values
were converted to CIE Lab color space. Then, histograms of each channel
in CIE Lab color space were calculated for all generated and reference
RAW images. Finally, the averaged HI was calculated as follows:

\begin{equation}
\mathrm{HI}=\frac{1}{3N}\sum_{c\in\left\{ L,a,b\right\} }\sum_{j=1}^{B}\min\left(\mathrm{H}_{ref,c}\left(j\right),\mathrm{H}_{gen,c}\left(j\right)\right),
\end{equation}
where $\mathrm{H}_{ref,c}$ and $\mathrm{H_{\mathit{gen,c}}}$ are
the histograms with $B$ bins for channel $c$ of all reference and
generated RAW images, $N$ denotes the number of total pixels, and
$\mathrm{H}_{ref,c}\left(j\right)$ and $\mathrm{H}_{gen,c}\left(j\right)$
are the number of pixels in the $j$-th bin. In this paper, we set
$B$ to $512$ and the range of values to $\left[-150,150\right]$.
This metric indicates how close the two marginal distributions of
pixels over all images are in terms of color and brightness.

All evaluation metrics reported in this paper are averaged values
of three-time experiments with different random seeds.

\subsection{RAW Image Object Detection}

CenterNet \cite{zhou2019objects} whose backbone is a modified DLA-34
\cite{yu2018deep} was used as a detector for eight class objects:
car, motorbike, bicycle, chair, diningtable, bottle, tvmonitor, and
bus \cite{Hong2021Crafting}. The detector was initialized with COCO
pretrained model and trained for 40 epochs with Adam \cite{kingma2014adam}
optimizer using the pseudo COCO-RAW images generated by each reversed
ISP method. A cosine decay learning rate schedule with a linear warmup
\cite{loshchilov2016sgdr} for the first 1,000 iterations whose maximum
and minimum learning rates were $10^{-3}$ and $10^{-4}$ was used.
The input size was (512, 512, 3), and random flip, random scaling,
and random cropping were used as data augmentation. In addition, we
also applied the RAW image noise injection proposed in \cite{Hong2021Crafting}
to the pseudo RAW images because the pseudo RAW images generated from
COCO dataset \cite{lin2014microsoft} contain less noises than real
RAW images of LOD dataset \cite{Hong2021Crafting}. Any color jitter
and cotrast augmentation were not used. We evaluated the detectors\textquoteright{}
performance with average precision (AP@0.5:0.95) \cite{lin2014microsoft}
for real LOD-RAW test images of three-time experiments with different
random seeds.

\section{Additional Results}

\subsection{Training Other Methods without Real-pairs}

To verify the importance of combining the proposed pseudo-pairs and
model architecture, we trained U-Net \cite{conde2022aim} and MBISPLD
\cite{conde2022model} with the same $\mathrm{PP_{\mathit{rand}}}$
as ours and evaluated them. Table \ref{tab:others_with_PP} shows
RAW image reconstruction results of ours and the other methods trained
without the real paired data on each dataset. Compared with the proposed
method, the other methods were not effectively trained using the $\mathrm{PP_{\mathit{rand}}}$,
which requires assuming multiple ISP pipelines. This is because it
was difficult to estimate the parameters of the forward ISP from the
input RGB image alone. As a result, in the training using all data,
only ours (all) was able to improve the accuracy greatly with the
benefit of the data amount. It shows that the proposed reference-guided
DPS plays an important role in the pseudo-pair training.

\subsection{Qualitative Results for RAW Image Datasets}

\begin{figure*}
\centering

\def\svgwidth{2.0\columnwidth}

\scriptsize
\begingroup%
  \makeatletter%
  \providecommand\color[2][]{%
    \errmessage{(Inkscape) Color is used for the text in Inkscape, but the package 'color.sty' is not loaded}%
    \renewcommand\color[2][]{}%
  }%
  \providecommand\transparent[1]{%
    \errmessage{(Inkscape) Transparency is used (non-zero) for the text in Inkscape, but the package 'transparent.sty' is not loaded}%
    \renewcommand\transparent[1]{}%
  }%
  \providecommand\rotatebox[2]{#2}%
  \newcommand*\fsize{\dimexpr\f@size pt\relax}%
  \newcommand*\lineheight[1]{\fontsize{\fsize}{#1\fsize}\selectfont}%
  \ifx\svgwidth\undefined%
    \setlength{\unitlength}{458.93115234bp}%
    \ifx\svgscale\undefined%
      \relax%
    \else%
      \setlength{\unitlength}{\unitlength * \real{\svgscale}}%
    \fi%
  \else%
    \setlength{\unitlength}{\svgwidth}%
  \fi%
  \global\let\svgwidth\undefined%
  \global\let\svgscale\undefined%
  \makeatother%
  \begin{picture}(1,0.68276986)%
    \lineheight{1}%
    \setlength\tabcolsep{0pt}%
    \put(0,0){\includegraphics[width=\unitlength,page=1]{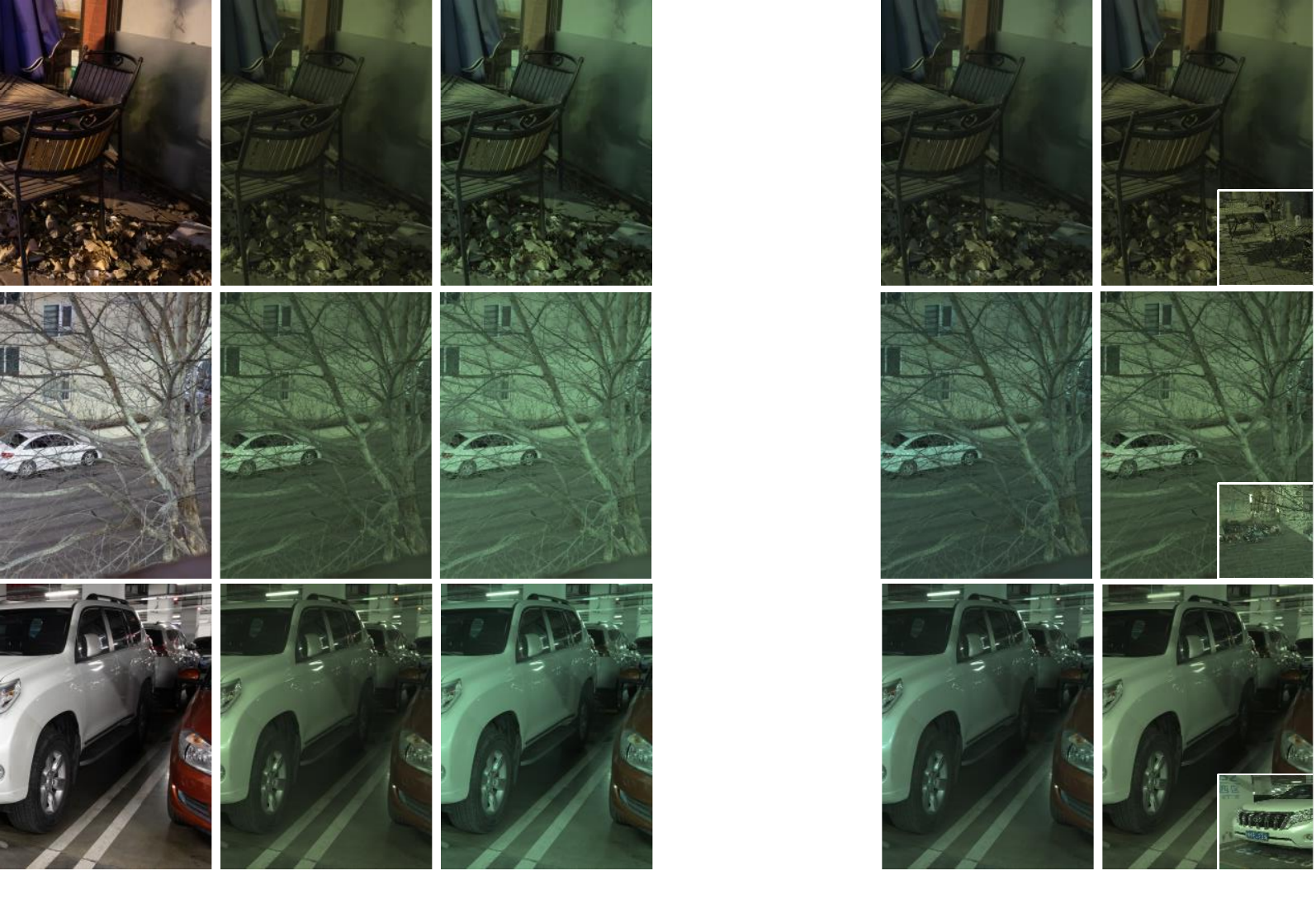}}%
    \put(0.04374512,0.00464145){\color[rgb]{0,0,0}\makebox(0,0)[lt]{\lineheight{1.25}\smash{\begin{tabular}[t]{l}Input RGB\end{tabular}}}}%
    \put(0.22149291,0.00464145){\color[rgb]{0,0,0}\makebox(0,0)[lt]{\lineheight{1.25}\smash{\begin{tabular}[t]{l}GT RAW\end{tabular}}}}%
    \put(0.72224339,0.00464145){\color[rgb]{0,0,0}\makebox(0,0)[lt]{\lineheight{1.25}\smash{\begin{tabular}[t]{l}MBISPLD \end{tabular}}}}%
    \put(0.87762617,0.00464145){\color[rgb]{0,0,0}\makebox(0,0)[lt]{\lineheight{1.25}\smash{\begin{tabular}[t]{l}Ours (COCO) \end{tabular}}}}%
    \put(0.40646619,0.00464145){\color[rgb]{0,0,0}\makebox(0,0)[lt]{\lineheight{1.25}\smash{\begin{tabular}[t]{l}UPI\end{tabular}}}}%
    \put(0,0){\includegraphics[width=\unitlength,page=2]{fig_results_lod_ppt.pdf}}%
    \put(0.56849484,0.00464145){\color[rgb]{0,0,0}\makebox(0,0)[lt]{\lineheight{1.25}\smash{\begin{tabular}[t]{l}U-Net\end{tabular}}}}%
  \end{picture}%
\endgroup%

\caption{Qualitative RAW reconstruction results for LOD Dataset. U-Net and
MBISPLD were trained with the real-pairs of LOD Dataset, and ours
(COCO) was trained with only the proposed pseudo-pairs ($\mathrm{PP_{\mathit{rand}}}$
and $\mathrm{PP_{\mathit{MT}}}$ for LOD and COCO dataset). The small
images on ours are the reference images. The gamma correction with
$\gamma=2.2$ was applied to all RAW images only for visualization.}

\label{fig:results_lod}
\end{figure*}

\begin{figure*}
\centering

\def\svgwidth{2.0\columnwidth}

\scriptsize
\begingroup%
  \makeatletter%
  \providecommand\color[2][]{%
    \errmessage{(Inkscape) Color is used for the text in Inkscape, but the package 'color.sty' is not loaded}%
    \renewcommand\color[2][]{}%
  }%
  \providecommand\transparent[1]{%
    \errmessage{(Inkscape) Transparency is used (non-zero) for the text in Inkscape, but the package 'transparent.sty' is not loaded}%
    \renewcommand\transparent[1]{}%
  }%
  \providecommand\rotatebox[2]{#2}%
  \newcommand*\fsize{\dimexpr\f@size pt\relax}%
  \newcommand*\lineheight[1]{\fontsize{\fsize}{#1\fsize}\selectfont}%
  \ifx\svgwidth\undefined%
    \setlength{\unitlength}{460.70498657bp}%
    \ifx\svgscale\undefined%
      \relax%
    \else%
      \setlength{\unitlength}{\unitlength * \real{\svgscale}}%
    \fi%
  \else%
    \setlength{\unitlength}{\svgwidth}%
  \fi%
  \global\let\svgwidth\undefined%
  \global\let\svgscale\undefined%
  \makeatother%
  \begin{picture}(1,1.11928485)%
    \lineheight{1}%
    \setlength\tabcolsep{0pt}%
    \put(0,0){\includegraphics[width=\unitlength,page=1]{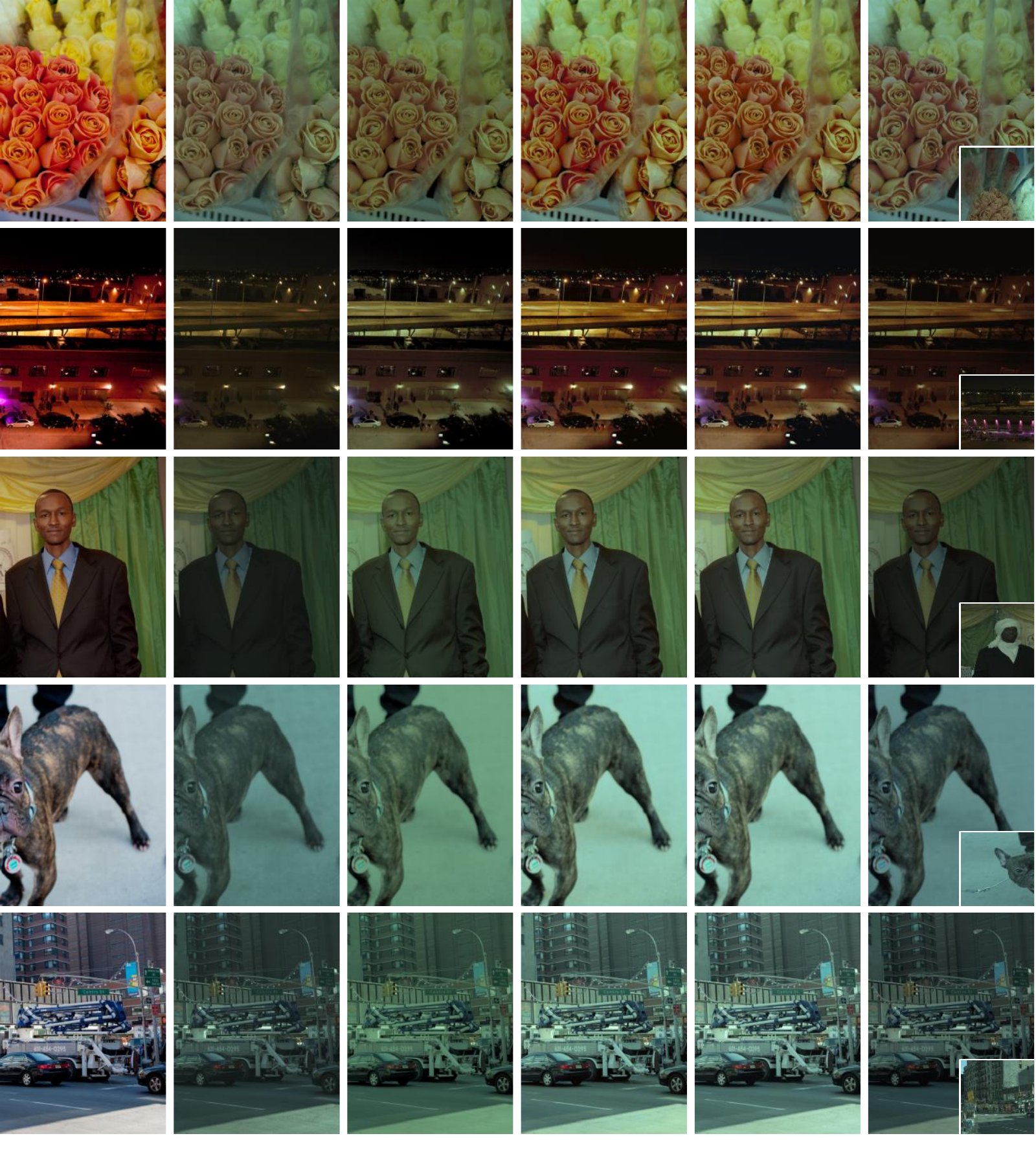}}%
    \put(0.04683257,0.00462361){\color[rgb]{0,0,0}\makebox(0,0)[lt]{\lineheight{1.25}\smash{\begin{tabular}[t]{l}Input RGB\end{tabular}}}}%
    \put(0.21738423,0.00462361){\color[rgb]{0,0,0}\makebox(0,0)[lt]{\lineheight{1.25}\smash{\begin{tabular}[t]{l}GT RAW\end{tabular}}}}%
    \put(0.71946257,0.00462361){\color[rgb]{0,0,0}\makebox(0,0)[lt]{\lineheight{1.25}\smash{\begin{tabular}[t]{l}MBISPLD \end{tabular}}}}%
    \put(0.87841461,0.00462361){\color[rgb]{0,0,0}\makebox(0,0)[lt]{\lineheight{1.25}\smash{\begin{tabular}[t]{l}Ours (Flickr) \end{tabular}}}}%
    \put(0.40490119,0.00462361){\color[rgb]{0,0,0}\makebox(0,0)[lt]{\lineheight{1.25}\smash{\begin{tabular}[t]{l}UPI\end{tabular}}}}%
    \put(0.56956187,0.00462361){\color[rgb]{0,0,0}\makebox(0,0)[lt]{\lineheight{1.25}\smash{\begin{tabular}[t]{l}U-Net\end{tabular}}}}%
  \end{picture}%
\endgroup%

\caption{Qualitative RAW reconstruction results for expert-tuned Nikon D700
(Expert C). U-Net and MBISPLD were trained with the real-pairs of
normal Nikon D700 (Libraw), and ours (Flickr) was trained with only
the proposed pseudo-pairs ($\mathrm{PP_{\mathit{rand}}}$ and $\mathrm{PP_{\mathit{MT}}}$
for normal Nikon D700 and Flickr dataset). The small images on ours
are the reference images. The gamma correction $\gamma=2.2$ was applied
to all RAW images only for visualization.}

\label{fig:results_nikon_c}
\end{figure*}

\begin{figure*}
\centering

\def\svgwidth{2.0\columnwidth}

\scriptsize
\begingroup%
  \makeatletter%
  \providecommand\color[2][]{%
    \errmessage{(Inkscape) Color is used for the text in Inkscape, but the package 'color.sty' is not loaded}%
    \renewcommand\color[2][]{}%
  }%
  \providecommand\transparent[1]{%
    \errmessage{(Inkscape) Transparency is used (non-zero) for the text in Inkscape, but the package 'transparent.sty' is not loaded}%
    \renewcommand\transparent[1]{}%
  }%
  \providecommand\rotatebox[2]{#2}%
  \newcommand*\fsize{\dimexpr\f@size pt\relax}%
  \newcommand*\lineheight[1]{\fontsize{\fsize}{#1\fsize}\selectfont}%
  \ifx\svgwidth\undefined%
    \setlength{\unitlength}{460.70498657bp}%
    \ifx\svgscale\undefined%
      \relax%
    \else%
      \setlength{\unitlength}{\unitlength * \real{\svgscale}}%
    \fi%
  \else%
    \setlength{\unitlength}{\svgwidth}%
  \fi%
  \global\let\svgwidth\undefined%
  \global\let\svgscale\undefined%
  \makeatother%
  \begin{picture}(1,1.12063457)%
    \lineheight{1}%
    \setlength\tabcolsep{0pt}%
    \put(0,0){\includegraphics[width=\unitlength,page=1]{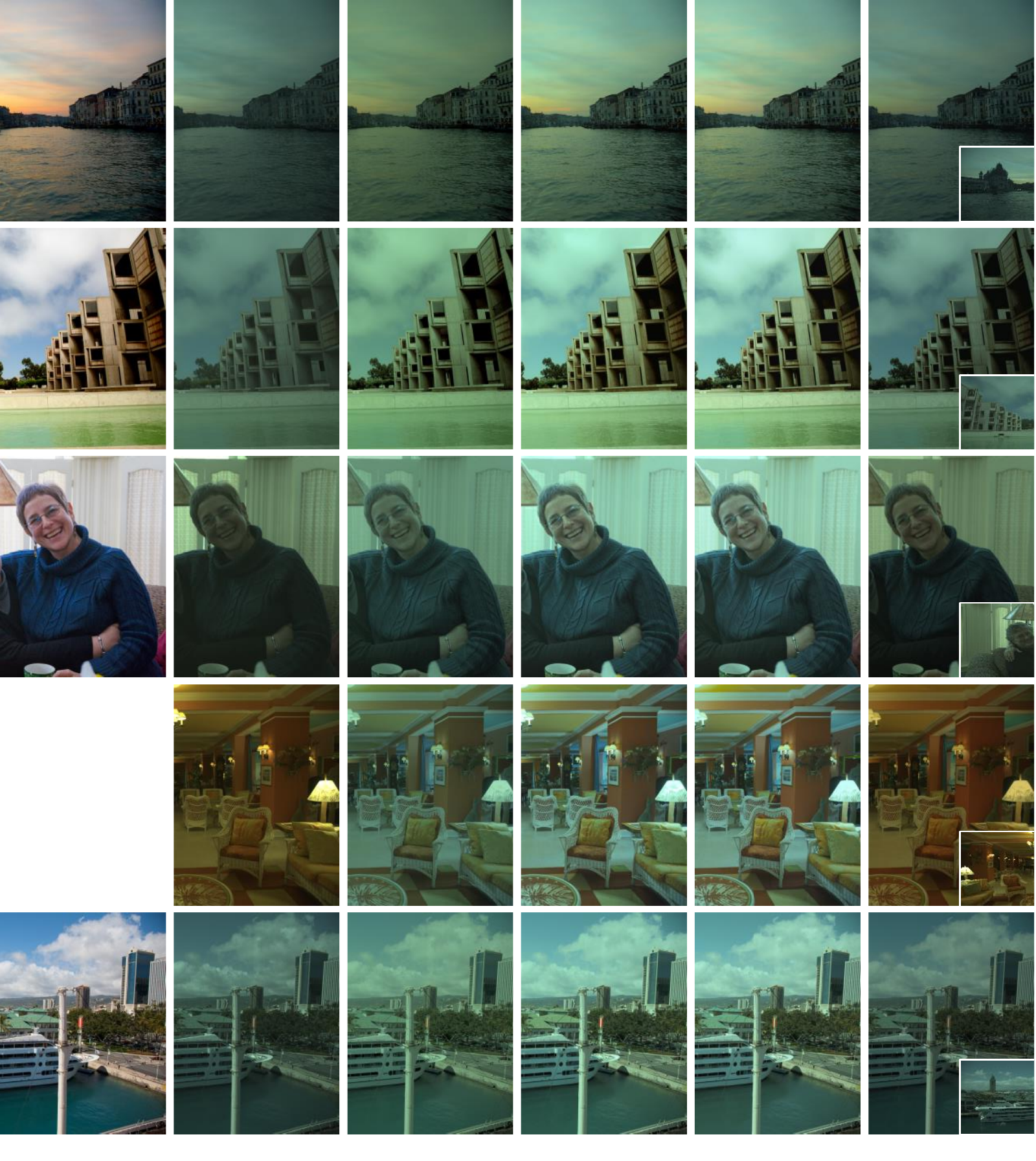}}%
    \put(0.04726453,0.0046236){\color[rgb]{0,0,0}\makebox(0,0)[lt]{\lineheight{1.25}\smash{\begin{tabular}[t]{l}Input RGB\end{tabular}}}}%
    \put(0.21781618,0.0046236){\color[rgb]{0,0,0}\makebox(0,0)[lt]{\lineheight{1.25}\smash{\begin{tabular}[t]{l}GT RAW\end{tabular}}}}%
    \put(0.71989452,0.0046236){\color[rgb]{0,0,0}\makebox(0,0)[lt]{\lineheight{1.25}\smash{\begin{tabular}[t]{l}MBISPLD \end{tabular}}}}%
    \put(0.87884656,0.0046236){\color[rgb]{0,0,0}\makebox(0,0)[lt]{\lineheight{1.25}\smash{\begin{tabular}[t]{l}Ours (Flickr) \end{tabular}}}}%
    \put(0.40533314,0.0046236){\color[rgb]{0,0,0}\makebox(0,0)[lt]{\lineheight{1.25}\smash{\begin{tabular}[t]{l}UPI\end{tabular}}}}%
    \put(0.56999383,0.0046236){\color[rgb]{0,0,0}\makebox(0,0)[lt]{\lineheight{1.25}\smash{\begin{tabular}[t]{l}U-Net\end{tabular}}}}%
    \put(0,0){\includegraphics[width=\unitlength,page=2]{fig_results_canon_c_ppt.pdf}}%
  \end{picture}%
\endgroup%

\caption{Qualitative RAW reconstruction results for expert-tuned Canon EOS
5D (Expert C). U-Net and MBISPLD were trained with the real-pairs
of normal Canon EOS 5D (Libraw), and ours (Flickr) was trained with
only the proposed pseudo-pairs ($\mathrm{PP_{\mathit{rand}}}$ and
$\mathrm{PP_{\mathit{MT}}}$ for normal Canon EOS 5D and Flickr dataset).
The small images on ours are the reference images. The gamma correction
with $\gamma=2.2$ was applied to all RAW images only for visualization.}

\label{fig:results_canon_c}
\end{figure*}

Figure \ref{fig:results_lod}, Figure \ref{fig:results_nikon_c},
and Figure \ref{fig:results_canon_c} show qualitative RAW reconstruction
results for LOD dataset, Nikon D700 (expert C tuning), and Canon EOS
5D (expert C tuning), respectively. Thanks to the reference guidance,
ours was able to generate more ground-truth-like images than the other
methods which were trained with the real image pairs. Although U-Net
and MBISPLD had the knowledge of the ISP pipeline of LOD dataset,
they failed to reconstruct the illumination color because there was
no clue in the input RGB image to estimate the color. However, our
method also has a issue in reconstructing highlight region. The bright
area of the generated images by ours tends to be darker than that
of the ground-truth RAW images. This issue may have been caused by
the design of the loss function, which focuses on reconstructing dark
areas. We suspect that the process of matching global brightness and
color was more emphasized than restoring highlight areas since the
pixels in a RAW image were essentially dark. The introduction of losses
that encourage the restoration of highlights and maintain the high
dynamic range is a future work.

\subsection{Qualitative Results for RGB Image Datasets}

\begin{figure*}
\centering

\def\svgwidth{2.0\columnwidth}

\scriptsize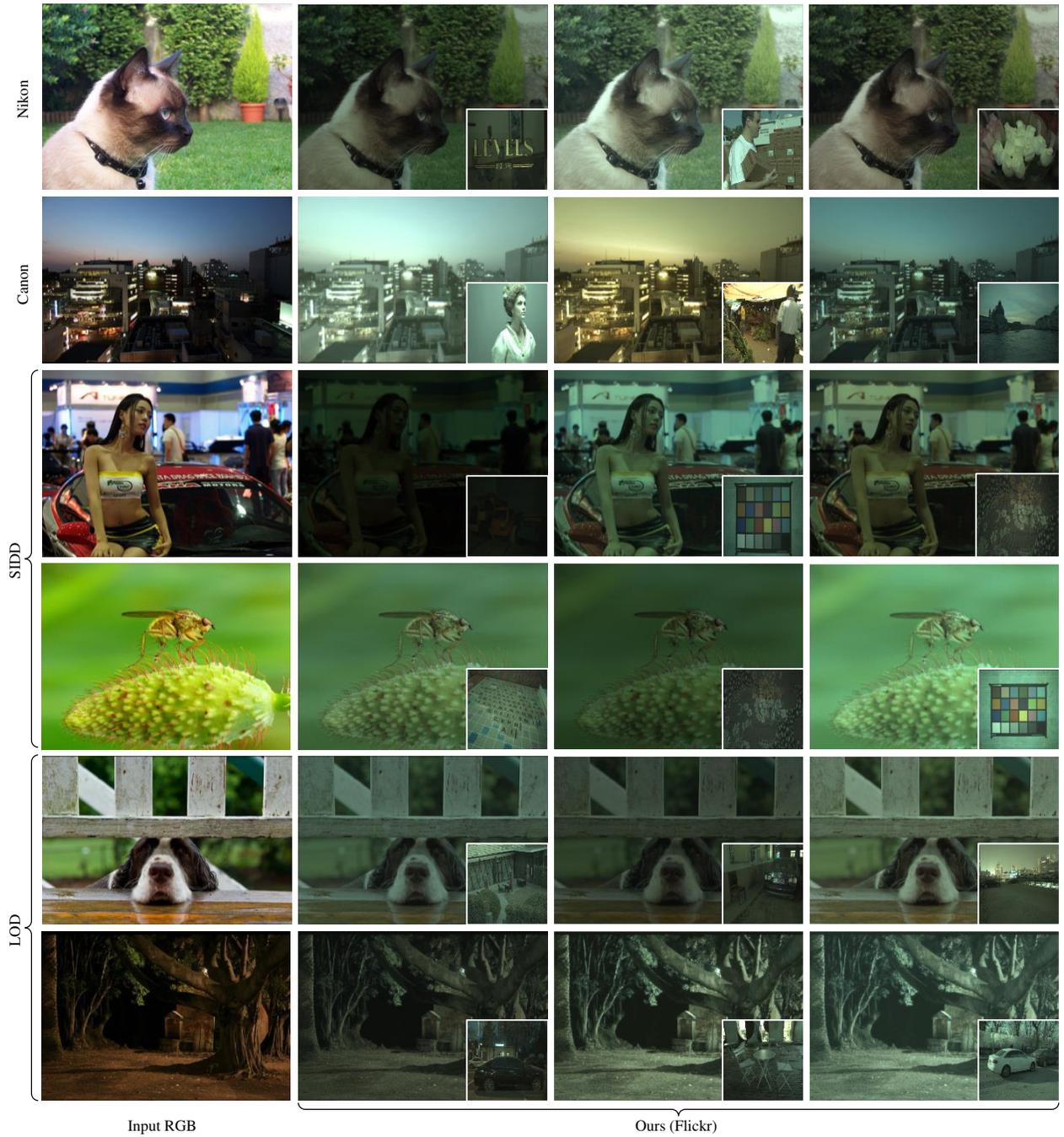

\caption{Qualitative RAW-like conversion results for RGB images in Flickr 1
Million Dataset. Ours (Flickr) was trained with only the proposed
pseudo-pairs for each RAW dataset (Nikon/Canon/SIDD/LOD dataset) and
Flickr dataset. The small images on ours are the reference images,
and the dataset name in the leftmost column denotes the name of the
dataset from which reference images were sampled. The gamma correction
with $\gamma=2.2$ was applied to all RAW images only for visualization.}

\label{fig:results_flickr}
\end{figure*}

\begin{figure*}
\centering

\def\svgwidth{2.0\columnwidth}

\scriptsize
\begingroup%
  \makeatletter%
  \providecommand\color[2][]{%
    \errmessage{(Inkscape) Color is used for the text in Inkscape, but the package 'color.sty' is not loaded}%
    \renewcommand\color[2][]{}%
  }%
  \providecommand\transparent[1]{%
    \errmessage{(Inkscape) Transparency is used (non-zero) for the text in Inkscape, but the package 'transparent.sty' is not loaded}%
    \renewcommand\transparent[1]{}%
  }%
  \providecommand\rotatebox[2]{#2}%
  \newcommand*\fsize{\dimexpr\f@size pt\relax}%
  \newcommand*\lineheight[1]{\fontsize{\fsize}{#1\fsize}\selectfont}%
  \ifx\svgwidth\undefined%
    \setlength{\unitlength}{509.43992615bp}%
    \ifx\svgscale\undefined%
      \relax%
    \else%
      \setlength{\unitlength}{\unitlength * \real{\svgscale}}%
    \fi%
  \else%
    \setlength{\unitlength}{\svgwidth}%
  \fi%
  \global\let\svgwidth\undefined%
  \global\let\svgscale\undefined%
  \makeatother%
  \begin{picture}(1,1.10149669)%
    \lineheight{1}%
    \setlength\tabcolsep{0pt}%
    \put(0.1207266,0.01092649){\color[rgb]{0,0,0}\makebox(0,0)[lt]{\lineheight{1.25}\smash{\begin{tabular}[t]{l}Input RGB\end{tabular}}}}%
    \put(0.59032465,0.01092649){\color[rgb]{0,0,0}\makebox(0,0)[lt]{\lineheight{1.25}\smash{\begin{tabular}[t]{l}Ours (COCO) \end{tabular}}}}%
    \put(0,0){\includegraphics[width=\unitlength,page=1]{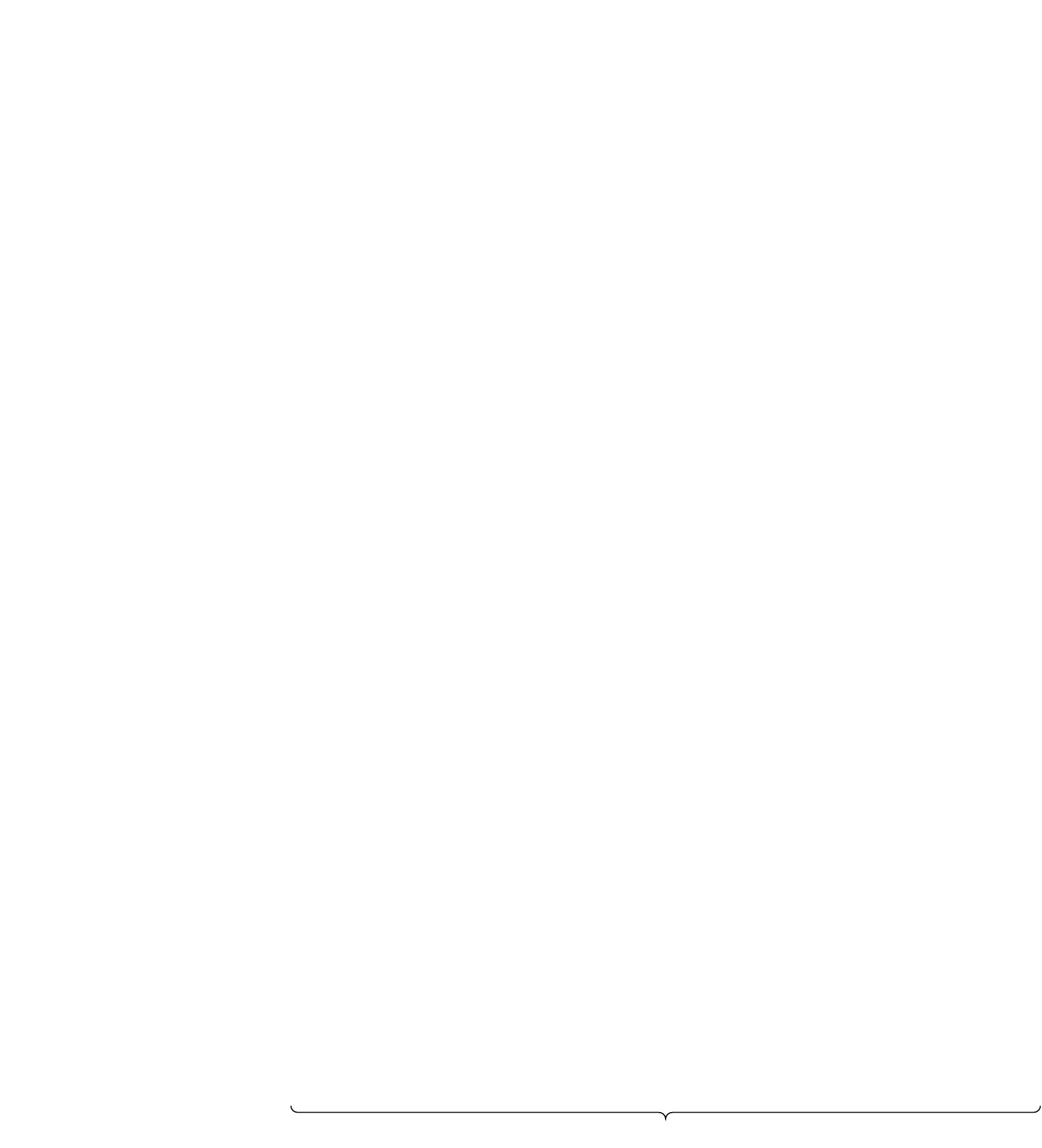}}%
    \put(0.01595272,0.20400126){\color[rgb]{0,0,0}\rotatebox{90.00000252}{\makebox(0,0)[lt]{\lineheight{1.25}\smash{\begin{tabular}[t]{l}LOD\end{tabular}}}}}%
    \put(0.01595272,0.56215928){\color[rgb]{0,0,0}\rotatebox{90.00000252}{\makebox(0,0)[lt]{\lineheight{1.25}\smash{\begin{tabular}[t]{l}SIDD\end{tabular}}}}}%
    \put(0.02339619,0.81575149){\color[rgb]{0,0,0}\rotatebox{90.00000252}{\makebox(0,0)[lt]{\lineheight{1.25}\smash{\begin{tabular}[t]{l}Canon\end{tabular}}}}}%
    \put(0.02339619,0.99153276){\color[rgb]{0,0,0}\rotatebox{90.00000252}{\makebox(0,0)[lt]{\lineheight{1.25}\smash{\begin{tabular}[t]{l}Nikon\end{tabular}}}}}%
    \put(0,0){\includegraphics[width=\unitlength,page=2]{fig_ours_coco_ppt.pdf}}%
  \end{picture}%
\endgroup%

\caption{Qualitative RAW-like conversion results for RGB images in COCO dataset.
Ours (COCO) was trained with only the proposed pseudo-pairs for each
RAW dataset (Nikon/Canon/SIDD/LOD dataset) and COCO dataset. The small
images on ours are the reference images, and the dataset name in the
leftmost column denotes the name of the dataset from which reference
images were sampled. The gamma correction with $\gamma=2.2$ was applied
to all RAW images only for visualization.}

\label{fig:results_coco}
\end{figure*}

Figure \ref{fig:results_flickr} and Figure \ref{fig:results_coco}
show RAW-like images converted from RGB images in Flickr or COCO dataset
by our methods. RAW images were sampled from RAW image datasets (Nikon/Canon/SIDD/LOD)
and given as reference images for the proposed method trained on each
RAW and RGB dataset. The proposed method was able to generate multiple
RAW-like images from one input RGB image by changing the reference
images. These results show that our method enables us to obtain diverse
RAW-like images with a combination of source RGB and target RAW images.

\end{document}